\pdfoutput=1
\documentclass{article}
\usepackage[T1]{fontenc}
\usepackage{iclr2027_conference,times}

\usepackage{amsmath,amsfonts,bm}

\def\eqref#1{equation~\ref{#1}}

\def\1{\bm{1}}

\DeclareMathAlphabet{\mathsfit}{\encodingdefault}{\sfdefault}{m}{sl}
\SetMathAlphabet{\mathsfit}{bold}{\encodingdefault}{\sfdefault}{bx}{n}

\usepackage{hyperref}
\usepackage{url}
\usepackage{booktabs}
\usepackage{amsmath}
\usepackage{graphicx}
\usepackage{xcolor}
\usepackage{algorithm}
\usepackage{algpseudocode}
\usepackage{multirow}
\usepackage{colortbl}
\definecolor{oursrow}{RGB}{232,245,233}
\definecolor{oursrowlite}{RGB}{243,250,243}
\definecolor{oraclerow}{RGB}{242,242,242}
\definecolor{defercell}{RGB}{214,230,252}
\definecolor{rejectcell}{RGB}{253,228,205}
\definecolor{correctcell}{RGB}{233,218,250}
\definecolor{offcell}{RGB}{224,224,224}

\iclrfinalcopy

\usepackage[skip=2pt plus1pt, font=small]{caption}
\makeatletter
\def\section{\@startsection {section}{1}{\z@}{-1.0ex plus
    -0.3ex minus -.2ex}{0.5ex plus 0.2ex minus 0.2ex}{\large\sc\raggedright}}
\def\subsection{\@startsection{subsection}{2}{\z@}{-0.85ex plus
    -0.3ex minus -.2ex}{0.3ex plus .2ex}{\normalsize\sc\raggedright}}
\def\subsubsection{\@startsection{subsubsection}{3}{\z@}{-0.9ex plus
    -0.4ex minus -.2ex}{0.3ex plus .2ex}{\normalsize\sc\raggedright}}
\makeatother
\AtBeginDocument{%
  \setlength{\abovedisplayskip}{4pt plus 1pt minus 2pt}%
  \setlength{\belowdisplayskip}{4pt plus 1pt minus 2pt}%
  \setlength{\abovedisplayshortskip}{2pt}%
  \setlength{\belowdisplayshortskip}{2pt}}

\newcommand{\sys}{\textsc{Gamma}}
\newcommand{\bank}{\mathcal{M}}
\newcommand{\sub}{\hat{g}}
\newcommand{\oraclesub}{g^{\ast}}

\title{\textsc{Gamma}: Grounded Multi-Agent Memory with\\ Mechanical Audit for Memory-Intensive Manipulation}

\author{Yudong Liu$^{1}$, Yueqian Lin$^{1}$, Zijia Tang$^{1}$, Yuxi Zheng$^{1}$, Yuan Li$^{2}$,\\
\textbf{Shuai Zhang$^{2}$, Yi Li$^{2}$, Taotao Jing$^{2}$, Shuangjun Liu$^{2}$, Dashan Gao$^{2}$,}\\
\textbf{Ning Bi$^{2}$, Jingwei Sun$^{3}$, Yiran Chen$^{1}$, Hai Helen Li$^{1}$}\\
$^{1}$Duke University \quad $^{2}$Qualcomm AI Research \quad $^{3}$University of Florida\\
{\fontsize{8.5}{10}\selectfont\texttt{\{yudong.liu,yueqian.lin,zijia.tang,yuxi.zheng,yiran.chen,hai.li\}@duke.edu}}\\
{\fontsize{8.5}{10}\selectfont\texttt{\{yuali,yli35,shuangju,shuazhan,tjing,dgao,nbi\}@qti.qualcomm.com}}\\
{\fontsize{8.5}{10}\selectfont\texttt{sun.jingwei@ufl.edu}}\\
Code: \url{https://github.com/1999Lyd/GAMMA}
}

\begin{document}

\maketitle
\lhead{Preprint. Under review.}

\begin{abstract}
Pretrained vision--language--action (VLA) policies fail on memory
tasks, behaviour that depends on events observed long before the
current frame, because their training data is overwhelmingly Markov and
mapping episodic memory directly to action chunks is hard. Yet the same policy, finetuned once to follow a privileged oracle's
grounded-subgoal text, nearly solves these tasks: the difficulty lies
entirely in mapping memory to a grounded symbolic subgoal. Existing work entrusts this stage to a
single VLM that selects keyframes and reasons over raw frames, and falls
far short of the oracle: key identities and dynamics are buried in
noisy pixels, the reasoning spans hundreds of frames in an unbounded
context, and hallucination corrupts the memory itself. We instead present \sys{}
(\textbf{G}rounded multi-\textbf{A}gent \textbf{M}emory with
\textbf{M}echanical \textbf{A}udit), which builds the stage as a
collaborating two-agent memory system: a detector tool
(SAM-3) for spatial grounding, a VLM \emph{writer} that distils each
replan window into one grounded text line, an append-only \emph{text
memory bank}, and a VLM \emph{reasoner} that reads the bank and emits
the next grounded subgoal at the oracle's cadence, wrapped in a
\emph{propose--verify harness} that admits, defers, corrects, or
rejects every agent claim against mechanical visual evidence, at
training-corpus construction and at deployment alike. On a sixteen-task
memory-manipulation suite \sys{} recovers a large fraction of the
oracle ceiling with the policy held fixed, reaching the ceiling on several
tasks, and transfers with unchanged contracts to a second benchmark.
\end{abstract}

\begin{figure}[t]
\centering
\includegraphics[width=0.85\linewidth]{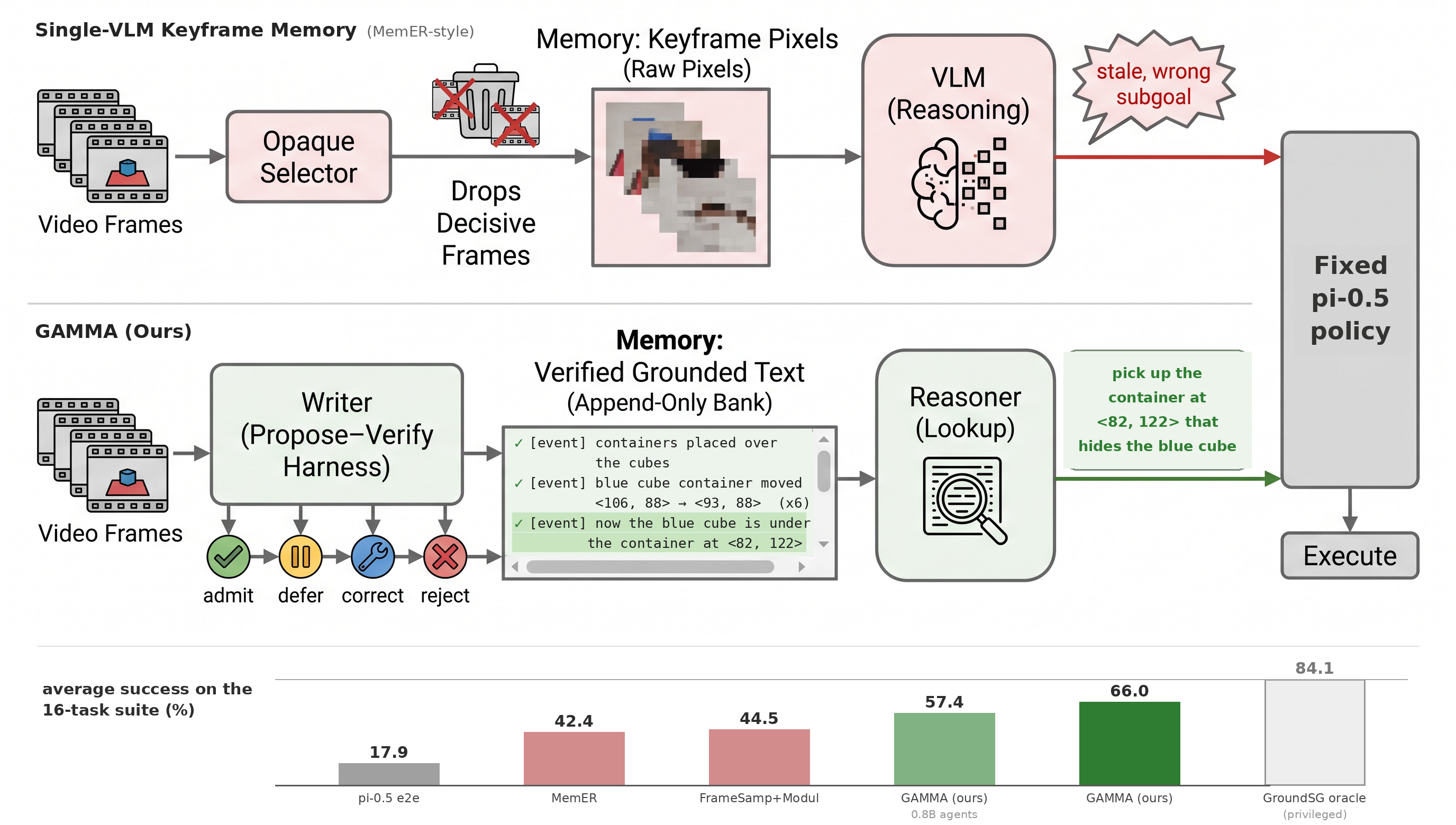}
\caption{The memory medium is the key difference.
\textbf{Top:} a MemER-style single-VLM keyframe pipeline remembers
pixels: an opaque selector drops decisive frames and the VLM
emits a stale, wrong subgoal. \textbf{Middle:} \sys{} remembers
verified grounded text: the propose--verify harness gates every
writer line into the append-only bank (lines from a real episode) and
the reasoner looks up exactly the line it needs. Both feed the
same fixed subgoal-conditioned $\pi_{0.5}$: what differs is
the memory system in between. \textbf{Bars:} suite-average success
vs.\ the privileged oracle.}
\label{fig:teaser}
\end{figure}

\section{Introduction}
\label{sec:intro}

Vision--language--action (VLA) policies pretrained on large manipulation
corpora~\citep{pi05,openvla,rt2} now solve a wide range of tabletop tasks
from a single language instruction. Their training data, however, is
overwhelmingly \emph{Markov}: the correct action chunk is a function of
the current observation and the instruction. Robot tasks in homes and
factories are frequently not of this kind. An agent that watched a
demonstration must reproduce it after the visual evidence is gone; an
agent that saw a container shuffle must remember which container now hides
which object; an agent asked to press a button after the fourth
pass of a moving object must count. On a recent memory-manipulation
benchmark~\citep{robomme} instantiating sixteen such tasks, the
$\pi_{0.5}$ policy~\citep{pi05}, finetuned end-to-end, succeeds on only
a small fraction of episodes (Table~\ref{tab:motivation}).

Two observations make this failure a well-posed problem. First,
finetuning the policy on the benchmark under the standard instruction
leaves success low even when a memory module injects past frames into
it ($0.179$ end-to-end, $0.445$ for the best memory-VLA architecture;
Table~\ref{tab:motivation}): the mapping from a long episodic history
to an action chunk lies far from the distribution the policy was
pretrained on. Second, the benchmark~\citep{robomme} ships the same
policy finetuned once to condition on a privileged oracle's
grounded-subgoal text, one grounded sentence per action chunk such as
\emph{``pick up the container at $\langle 105, 87\rangle$ that hides
the green cube''}, and it nearly solves the suite ($0.841$): a
grounded subgoal returns the data to the pretraining distribution,
Markov, simple and direct, since the behaviour is again a function of
the current observation and one instruction-like sentence.
Subgoal-to-action is thus solved by the pretrained VLA's own low-level
control; learning memory-to-action directly would couple that control
with high-level memory reasoning, the hard part, whereas decomposing
the problem as
\[
\text{memory} \;\longrightarrow\; \text{grounded symbolic subgoal}
\;\longrightarrow\; \text{action chunk}
\]
isolates the genuinely unsolved stage and inherits the policy's motor
competence without further training. The problem is therefore to
produce the oracle's subgoal stream (content, coordinates, and
timing) from raw episodic observation.

The natural candidate for this first stage is a single VLM over the
episode's frames, as in keyframe-memory pipelines such as
MemER~\citep{memer}: select a small set of informative frames, then reason
over them to produce the next subgoal. Benchmark results show that this family
falls well short of the oracle
(Figure~\ref{fig:teaser}), and our tick-level analysis
(Section~\ref{sec:diagnosis}) explains why. (i)~\emph{Keyframe selection
loses information}: nothing guarantees the decisive frame survives
selection, the selector's policy is opaque, and a dropped frame is
unrecoverable. (ii)~\emph{VLMs are weak at precise spatial grounding}:
grounded subgoals carry exact coordinates, and a $\langle x,y\rangle$
off by a few pixels sends the policy to the wrong one of $N$ identical
objects. (iii)~\emph{Reasoning over raw frames is hard}: the key object's
identity and motion are buried in redundant background, and the
frame-token context grows without bound. (iv)~\emph{VLMs hallucinate}:
a VLM asserts what is plausible rather than what is
observed, uniquely damaging in a memory system, where one false fact
enters the record and poisons every decision downstream. The first three
failures share a root---\emph{pixels are the wrong medium for episodic
memory}---and the fourth demands that whatever medium replaces them be
verifiable.

\begin{figure}[t]
\centering
\includegraphics[width=0.68\linewidth]{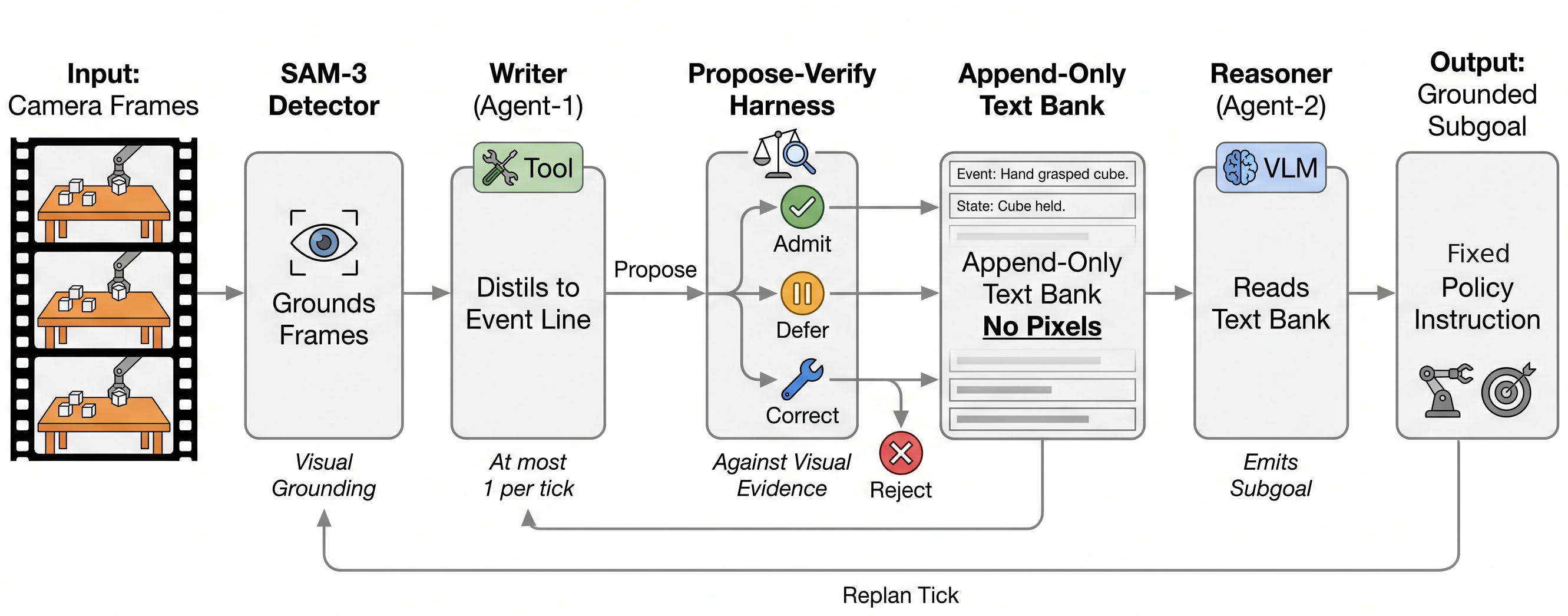}
\caption{Overview of \sys{}. Each replan tick, the SAM-3 detector
grounds the window's frames; the writer (Agent-1) distils them into at
most one grounded event line; the reasoner (Agent-2) reads the
append-only bank, without pixels, and emits the next grounded subgoal.
The agents' only interface is the bank; every claim must pass the
propose--verify harness against visual evidence.}
\label{fig:system}
\end{figure}

We therefore build the memory stage as \sys{}, a grounded
multi-agent system rather than a single model
(Figure~\ref{fig:system}). A segmentation-based detector
(SAM-3~\citep{sam3}), called as a tool on every frame, supplies the
precise coordinates a VLM produces unreliably. A VLM writer (agent-1)
runs once per replan tick and distils its window of sampled frames,
their detections, the memory so far and the planner's expected next
subgoal into at most one grounded text line, or none when the window
shows no task-relevant change. The append-only text memory bank is the
only history carrier: faithful (every line is evidence-gated against
its own frames), compact (hundreds of steps become a handful of lines)
and transparent (every coordinate in it came from the detector). A VLM
reasoner (agent-2) reads the bank alone, without pixels, and emits the
next grounded subgoal, trained to match the oracle stream verbatim at
its cadence. Grounded text alone does not remove failure (iv): in
closed loop the finetuned writer completed actions on a learned
schedule rather than on evidence, and both agents cited coordinates
they could not know (Section~\ref{sec:diagnosis}). Because the bank is
symbolic and grounded, every such claim is checkable, and we therefore
wrap both agents in a propose--verify harness
(Section~\ref{sec:harness}, Algorithm~\ref{alg:harness}): every claim
the agents make (an event line proposed for memory, a completion that
would advance the plan, a cited coordinate) is checked against
mechanical visual evidence and admitted, deferred until its evidence
appears, corrected to the evidence-consistent value, or rejected. The
same contract generates the supervision (Section~\ref{sec:targets}): a
target line exists only at the window that shows its evidence and
cites only coordinates visible in the writer's inputs, so the agents
are trained on the verified distribution the harness enforces at
deployment; the reasoner is further trained on banks written by the
trained writer rather than on ground-truth banks, so its training
distribution matches deployment.

Across the full suite, \sys{} recovers a large fraction of the
oracle ceiling with the policy fixed ($0.660$ average success against
the oracle's $0.841$, $78.5\%$ of the privileged ceiling,
Table~\ref{tab:main}), reaching it on several tasks. Because the memory
is readable text, every failure can be attributed to a bank line or its
absence, yielding design principles validated in closed loop
(Section~\ref{sec:diagnosis}, Appendix~\ref{app:rounds}).

\textbf{Contributions.}
(1)~A decomposition argument, backed by a controlled end-to-end-versus-
oracle gap under one fixed policy, locating the difficulty of VLA memory
tasks in the memory-to-grounded-subgoal stage.
(2)~\sys{}, a grounded multi-agent memory system (detector tool, VLM
writer, append-only text bank, VLM reasoner) that replaces opaque
keyframe selection with verified symbolic memory, and whose
propose--verify harness is a general contract under which every agent
claim (event, progress or grounding) takes effect only after a
mechanical evidence check that admits, defers, corrects or rejects it,
applied identically at corpus construction and at deployment.
(3)~State-of-the-art memory-task performance with a fixed policy on
RoboMME, transfer with unchanged contracts to a second benchmark, and a
tick-level diagnostic protocol that turns closed-loop failures into
attributable, fixable memory defects.

\section{Related Work}
\label{sec:related}

\textbf{Vision--language--action policies.} Imitation policies for
manipulation~\citep{act,diffusionpolicy} and their scaled,
language-conditioned successors~\citep{rt1,rt2,palme,octo,openvla,
openvlaoft,robomamba,pi0,pi05} map the current observation and an
instruction to short action chunks, from training data dominated by
tasks fully observable in the current
frame~\citep{robomimic,contextvla}. We treat the fixed
subgoal-conditioned policy (Section~\ref{sec:problem}) as a competent
executor and study the memory stage above it.

\textbf{Memory for embodied agents.} Partial observability in control
has classically been met with \emph{architectural}
memory~\citep{drqn,merlin,gtrxl,amago,rmt,ttt,elmur}: representations
that are latent, and so cannot be inspected, audited, or corrected at
deployment. In VLAs, recent work compresses past frames into the policy's
context~\citep{contextvla} or attaches a memory module to the
backbone or recurrent queries~\citep{memoryvla,rememvla}; a growing 2026
line gives VLAs explicit
memory~\citep{unimemvla,eventvla,weavela,aura,agm,hime}, audits show frozen
VLAs barely deploy the history they encode~\citep{pbnr}, and a
contrasting line argues for continuous latent reasoning~\citep{crvla}. Closest to us,
MemER~\citep{memer} keeps an episodic memory of VLM-selected keyframes
and reasons over the selection with a second VLM pass. We share the
decompose-to-subgoal philosophy but replace pixel memory with verified
grounded text and selection with per-tick writing: nothing is
discarded, and every remembered fact is a checkable sentence.

\textbf{Benchmarks.} Standard manipulation
suites~\citep{metaworld,rlbench,calvin,libero,robocasa} evaluate
long-horizon chains whose sub-tasks are individually Markov, so strong
policies need no episodic store. Recent benchmarks target the missing
dimension~\citep{mikasarobo,robocerebra,rmbench}; RoboMME~\citep{robomme}
isolates episodic memory itself: sixteen tasks built around counts,
orders, swaps, and demonstrated routes, with a privileged
grounded-subgoal oracle that upper-bounds subgoal-conditioned
execution. RoboMemArena~\citep{rma} adds 26 long-horizon kitchen
tasks in four memory families. We adopt RoboMME's tasks, success
predicates, and oracle cadence of one subgoal per action chunk, and
evaluate on both.

\textbf{VLM agents with tools and verification.} LLM planners over
robot skills~\citep{saycan,codeaspolicies} choose the next step from
language; Inner Monologue~\citep{innermonologue} showed that textual
observations fed back into the loop make it robust, a precursor of our
text bank, and append-only natural-language memories underpin
open-ended agents~\citep{generativeagents,voyager}. Like tool-using
agents~\citep{toolformer,react} and visual
programs~\citep{visprog,vipergpt}, our system delegates what the model
does poorly: SAM-3~\citep{sam3} supplies pixel-exact grounding, and
evidence extractors the percepts sparse-frame VLM inference reads at
chance. Verification of model output ranges from
self-critique~\citep{reflexion,selfrefine} to trained process
verifiers~\citep{verifystep}, model judgments of model text; the
harness instead admits a claim by a mechanical check against perception.

\section{Problem Formulation}
\label{sec:problem}

\textbf{Setting.} Each episode of a memory task presents an instruction $I$, a
\emph{demonstration phase} (a rendered video prefix showing the required
behaviour, e.g.\ a route to trace or an order to reproduce), and an
\emph{execution phase} in which the robot acts. An episode is scored by
the environment's success predicate. The sixteen tasks cover counted
actions (press a button after the $N$-th pass), object permanence under
occlusion and shuffling (which container hides which cube), demonstrated
orders and routes, and delayed reveals; in all of them the information
needed at execution time was observable earlier and is absent from the
current frame.

\textbf{Oracle subgoal stream.} The benchmark's privileged planner emits,
at every tick $k$ (one tick $=$ one action chunk of $H$ control steps,
the policy's replan cadence), a grounded subgoal string $\oraclesub_k$: one sentence naming
the current sub-task and, when it refers to an object, that object's
current image coordinate. $\pi_{0.5}$ finetuned to condition on
$\{\oraclesub_k\}$ attains $0.841$ success; finetuned end-to-end
without subgoals it attains $0.179$ (Table~\ref{tab:motivation}; all
five numbers are RoboMME's reported averages); the subgoal-conditioned
policy is trained once, on the oracle stream only, and never modified
afterwards.

\textbf{Goal.} Learn a system $F$ that, from the observable episode
alone, emits subgoals $\sub_k = F(I, o_{0:Hk})$, where $o_{0:t}$
denotes the stream of camera frames (demonstration and execution) up to
control step $t$, such that feeding $\{\sub_k\}$ to the fixed
subgoal-conditioned policy recovers the oracle's success rate. The
target is exacting in three ways: \emph{content} (the right
sub-task), \emph{grounding} (coordinates within pixels of the true
object among distractors), and \emph{timing} (the switch must
occur at the tick the previous sub-task completes; an early switch
asks the policy to skip undone work, a late one stalls it).

\begin{table}[t]
\centering
\footnotesize
\begin{tabular}{lcc}
\toprule
Configuration & Memory mechanism & Success \\
\midrule
$\pi_{0.5}$ end-to-end & none & $0.179$ \\
$\pi_{0.5}$ + memory module (FrameSamp+Modul) & latent frame memory & $0.445$ \\
MemER-style keyframe pipeline & VLM keyframe selection & $0.424$ \\
$\pi_{0.5}$ + oracle \emph{ungrounded} subgoals & privileged, no coordinates & $0.496$ \\
$\pi_{0.5}$ + oracle grounded subgoals & privileged & $0.841$ \\
\bottomrule
\end{tabular}
\caption{The policy is not the bottleneck, and grounding sets the
ceiling: oracle grounded subgoals take $\pi_{0.5}$ from $0.179$ to
$0.841$; the same oracle without coordinates reaches $0.496$, memory
injected into the policy $0.445$. Our task is to supply the grounded
stream without privilege (all rows RoboMME-reported).}
\label{tab:motivation}
\end{table}

\section{Method}
\label{sec:method}

\subsection{System structure}
\label{sec:system}

\sys{} runs on the tick grid of the policy: both agents execute once
per $H$-step action chunk, writer first, over a window of $n$ uniformly
spaced frames (all constants in Appendix~\ref{app:config}). The agents
share no hidden state: they collaborate only through the bank, so the
entire inter-agent interface is readable, auditable text.

\textbf{Detector (grounding tool).} SAM-3 with per-class tuned text
prompts detects the task-relevant object classes (cubes by colour,
containers, buttons, targets, bins, pegs, sticks, the arm) in every
window frame, yielding per-frame detection strings
$D_t = \{\,\text{name}\langle x, y\rangle\,\}$, each entry naming an
object class and its image coordinate $\langle x, y\rangle$ in frame
$o_t$. The identical detector,
prompts, and post-processing run at training-data generation and at
deployment, so the writer's input channel is reproduced verbatim, a
train/serve contract we found essential (Appendix~\ref{app:rounds}).

\textbf{Agent-1: the memory writer.} What memory reasoning needs from
perception is not pixels but two kinds of facts: \emph{visual identity}
$\iota$, binding an object's task role (the container hiding a
particular cube, the demonstrated target) to a detector coordinate,
across occlusion and among identical instances; and \emph{visual
dynamics} $\delta$, the change the window shows (an object appearing,
vanishing, or moving; an action starting or completing). Everything the
reasoner will later need is an identity--dynamics pair; everything else
in the frames is noise. The writer $A_1$, a finetuned VLM, extracts
exactly this pair from its window, serialised as the bank line $m_k$:
\begin{equation}
m_k \;=\; (\iota_k, \delta_k) \;=\; A_1\!\left(I,\; o_{H(k-1):Hk},\;
D_{H(k-1):Hk},\; \bank_{k-1},\; \sub_{k-1}\right),
\end{equation}
where $\bank_{k-1}$ is the memory bank after tick $k{-}1$ (the
append-only list of admitted text lines formalised below) and
$\sub_{k-1}$ is the reasoner's previous subgoal. The line $m_k$ is at
most one declarative sentence naming identities by role with
coordinates and the dynamic relating them, or \textsc{none} if the
window holds no new fact. Tick 0 writes the initial-scene line listing
every detected object.

\textbf{Text memory bank.} The bank $\bank$ is an append-only list,
$\bank_k = \bank_{k-1} \oplus m_k$ (where $\oplus$ appends the admitted
line), never rewritten or summarised. It is
the only state carried between ticks. Append-only text gives three
properties keyframe memories lack: no selector destroys information;
every remembered fact is checkable against the frames that produced it;
and hundreds of steps compress to a handful of lines, so the full
episode always fits in a small context.

\textbf{Agent-2: the reasoner.} The reasoner $A_2$, a finetuned VLM
sharing the writer's base, reads the instruction and the entire
bank, with no pixels at all, and emits the next subgoal:
\begin{equation}
\sub_k \;=\; A_2\!\left(I,\; \bank_k\right),
\end{equation}
trained so that $\sub_k = \oraclesub_k$ verbatim, including the phase
conventions (\emph{``hold and wait for the demo to complete''} during the
demonstration; \emph{``all tasks completed; remain static''} at the
end). The reasoner needs no pixels: giving it the current frame did not
help in closed loop (Appendix~\ref{app:rounds}); progress judgment belongs to
the writer and the harness, and fresh coordinates reach it through
admitted bank lines. It may only cite coordinates that exist in the
bank; a grounded subgoal the bank cannot yet support is replaced in
training by \emph{``hold and observe the scene''}. The policy then
executes $a_{Hk:H(k+1)} = \pi_{0.5}(o_{Hk}, \sub_k)$ with exactly the
oracle row's weights, never trained on our outputs.

\subsection{The propose--verify harness}
\label{sec:harness}

\begin{figure}[t]
\centering
\begin{minipage}[t]{0.468\linewidth}
\includegraphics[width=\linewidth]{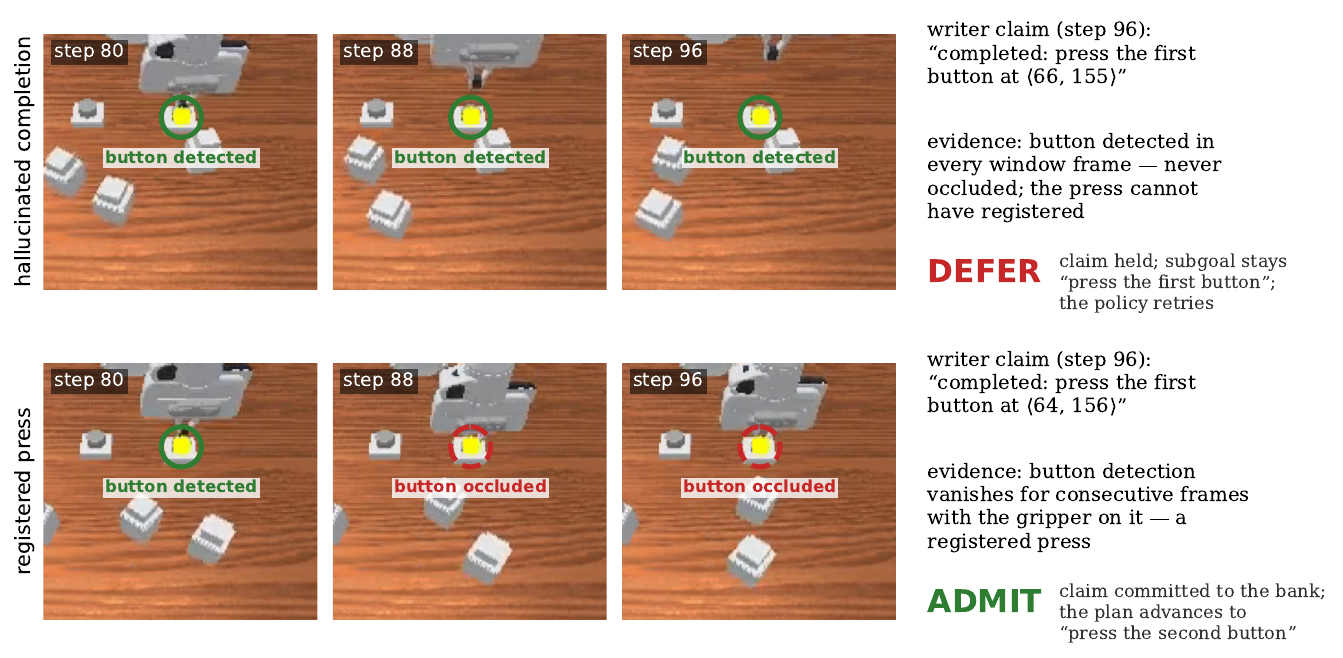}
\end{minipage}\hfill
\begin{minipage}[t]{0.515\linewidth}
\includegraphics[width=\linewidth]{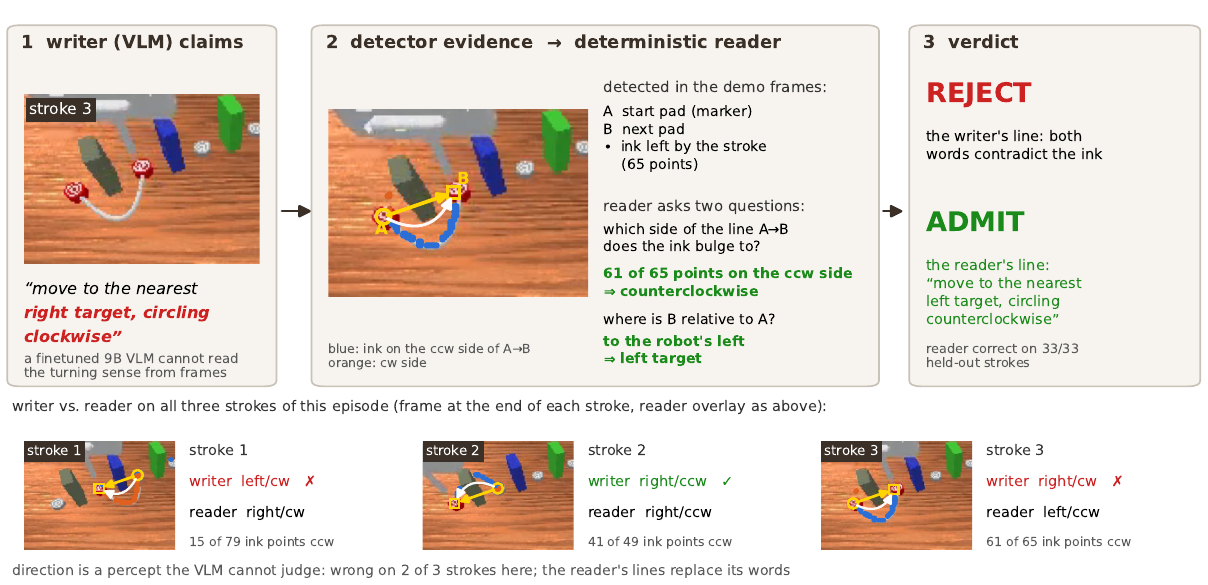}
\end{minipage}
\caption{The harness on real claims. \textbf{Left}
(ButtonUnmaskSwap), the claim ``completed: press the first button'': the
button surviving every window frame refutes it---\textsc{defer}, and the
policy retries (top); its disappearance under the gripper registers the
press---\textsc{admit} (bottom). \textbf{Right} (RouteStick, stroke 3):
the writer says ``right target, clockwise''; from the start pad $A$, the
next pad $B$ and the stroke's ink, the reader finds the ink bulging
counterclockwise (61 of 65 points) with $B$ to the left. Both words are
contradicted: the writer's line is \textsc{reject}ed, the reader's
admitted.}
\label{fig:harness-cases}
\end{figure}

The agents are fluent proposers, not reliable reporters: absent any
check, the writer sometimes completes invisible actions on its learned
schedule rather than on evidence, and both agents occasionally cite
coordinates they cannot know. The harness makes unreliability harmless by interposing a fixed
contract between proposal and effect.

\textbf{Claims.} Everything the agents emit is a set of typed
\emph{claims} $\mathcal{C}_k$ at tick $k$: \emph{event claims} (a line
proposed for the bank), \emph{progress claims} (a completion that would
license the plan's transition), and \emph{grounding claims} (every
cited coordinate binding a role to a location).

\textbf{Verdicts.} A verifier evaluates each claim against the tick's
evidence $e_k$ (the window's frames and detections) with a per-type
predicate $V(c, e_k) \in
\{\textsc{admit}, \textsc{defer}, \textsc{correct}(c'),
\textsc{reject}\}$:
\textsc{admit} lets the claim take effect; \textsc{defer} holds it in a
pending set with no effect, so the previous subgoal keeps being issued,
turning a premature transition into a retry, and the claim is
re-checked each tick until a terminal verdict or the end of the episode
(Figure~\ref{fig:harness-cases}, left): a deferral never stalls the
policy, only the transition, and missing evidence ends the episode as a
failure at its step limit rather than with a false step (cost measured
by the \textsc{defer} ablation, Section~\ref{sec:ablation});
\textsc{correct} commits a repaired copy $c'$ whose grounding is
rebound to the evidence-consistent value (the nearest class-compatible
detection, or the fold of the writer's own admitted chain);
\textsc{reject} discards a claim outright: a duplicate, or a claim
type owned by an authoritative evidence extractor (the limiting case
below; Figure~\ref{fig:harness-cases}, right). The predicates use a
small vocabulary of generic percepts (presence or absence of a
detection at a location, occlusion, arrival within a radius,
displacement above a threshold), parameterised per claim schema, never
per episode (Appendix~\ref{app:motion}).

\textbf{One contract, both phases.} The same verifier generates the
training corpus (Section~\ref{sec:targets}): a target line is emitted
only at the window whose evidence admits it, and grounding that fails
verification is demoted rather than fabricated, so the agents learn the
verified distribution and the harness enforces it under the
distribution shift of deployment. In the limiting
case, a claim type whose evidence the proposer measurably reads at chance
(continuous stroke motion; Appendix~\ref{app:motion}), the verifier's
evidence extractor supplies the claim itself, and proposal and
verification coincide. \sys{} is thus a hybrid learned--programmed
system: on two of the sixteen tasks the demonstration claims come from
deterministic readers rather than the writer
(Appendix~\ref{app:motion}).

\begin{algorithm}[tb]
\caption{Propose--verify harness (one tick, generalised)}
\label{alg:harness}
\scriptsize
\begin{algorithmic}[1]
\Require claims $\mathcal{C}_k$ from the agents; evidence $e_k$;
pending set $\mathcal{P}$; verified progress index $j$
\For{each claim $c \in \mathcal{P} \cup \mathcal{C}_k$}
  \State $v \gets V_{\mathrm{type}(c)}(c, e_k)$;\;
  \textbf{if} $v \neq \textsc{defer}$ \textbf{then}
  $\mathcal{P} \gets \mathcal{P} \setminus \{c\}$
  \Comment{terminal verdicts clear pending}
  \If{$v = \textsc{admit}$}
     \State commit $c$ to $\bank$;\; \textbf{if} $c$ is a progress claim
     \textbf{then} $j \gets j + 1$
  \ElsIf{$v = \textsc{correct}(c')$}
     \State commit $c'$ to $\bank$ \Comment{grounding rebound to evidence}
  \ElsIf{$v = \textsc{defer}$}
     \State $\mathcal{P} \gets \mathcal{P} \cup \{c\}$
     \Comment{no effect; retried next tick; never evicted}
  \Else
     \State discard $c$ \Comment{reject: duplicate, or owned by an extractor}
  \EndIf
\EndFor
\State issue $\sub_k$ conditioned on the verified bank; when the plan is
a recorded sequence, pin $\sub_k$ to its $j$-th step
\Comment{deferred progress $\Rightarrow$ the policy retries}
\end{algorithmic}
\end{algorithm}

\subsection{Targets and training}
\label{sec:targets}

Both agents are finetuned on a corpus generated mechanically from the
benchmark's recorded episodes, the train-time counterpart of the harness. A
writer target is a line of a small, task-agnostic schema: a
\emph{kind} (scene, demonstration step, observation, completion) and one
sentence that names identities by role and grounds them with detector
coordinates, or \textsc{none}; each line is attached to the window
whose frames evidence it (Appendix~\ref{app:targets}). The generator
enforces three contracts on them.

\textbf{Event lines from verified trackers.} Writer targets are derived
from the oracle's subgoal boundaries (\emph{completed:} lines at the tick
the online stream switches; \emph{demo showed:} lines during the
demonstration), augmented with observation lines for events the benchmark
never annotates but the bank must carry: covering events, container-move
chains during shuffles (\emph{``the red cube's container moved from
$\ldots$ to $\ldots$''} with real detected endpoints), highlight
appearances, and per-stroke demo records. Augmented lines come from offline trackers; the oracle's role for
these lines is verification and correction. More broadly, both agents
are behaviour-cloned from the privileged planner's stream on training
episodes (the oracle is the teacher), while all evaluation episodes
are held out.

\textbf{Evidence gates.} A line may only be emitted at a tick whose own
window frames contain its evidence: an \emph{appeared} line requires the
object to be newly present, a \emph{move} line requires a measured
displacement, a \emph{completed} line requires visible motion. On the
released corpus both gates pass at near-unity rates
(Appendix~\ref{app:bankquality}). This rule makes annotation-timing
noise harmless (an event moves to the window that shows it) and is
why the writer can be trained to be faithful rather than fluent.

\textbf{Knowability.} Every coordinate in a target must be findable in
the writer's visible inputs (the window's detections or the bank lines
it sees). Lines whose coordinates fail this test are demoted to their
coordinate-free form. Without this rule the writer learns to emit
plausible-looking coordinates it cannot know, exactly the failure that
sends the policy to the wrong one of $N$ identical objects.

The reasoner's target at tick $k$ is the oracle string current at
$k{+}1$, coordinates snapped to the bank; during the demonstration
the hold sentence, after annotation exhaustion the terminal sentence.

\textbf{Two-stage training.}\label{sec:training} Both agents share one
base VLM (Qwen3.5-9B~\citep{qwen35}) with LoRA adapters~\citep{lora}.
The writer is finetuned first, on the verified event-line corpus with
ground-truth banks in context; it is then rolled out over the training
set, and the reasoner is finetuned on the resulting predicted banks with
oracle subgoals as targets, so it learns to plan from the banks it will
actually receive, including the writer's characteristic omissions and
phrasings. We train a second pair of agents by the same procedure on a
Qwen3.5-0.8B base (same corpus, contracts, detector, executor, only
the base model changed); both pairs are evaluated
(Section~\ref{sec:ablation}) and released.

Deployment reproduces the training geometry: the
demonstration prefix is replayed through the writer to build the demo
record, then writer and reasoner run once per tick behind the harness
(Appendix~\ref{app:config}); all admitted content is model-written
except the limiting-case claims of Appendix~\ref{app:motion}.

\section{Experiments}
\label{sec:experiments}

\subsection{Setup}
\textbf{Benchmarks.} RoboMME (sixteen memory tasks, held-out test
episodes); RoboMemArena (26 tasks, held-out layouts; Section~\ref{sec:rma}).
\textbf{Models.} Writer and reasoner are LoRA adapters on one
Qwen3.5-9B base, trained in two stages as in
Section~\ref{sec:targets} (hyperparameters in
Appendix~\ref{app:config}); a second pair is trained identically on a
Qwen3.5-0.8B base and is the scale axis of
Section~\ref{sec:ablation}. The detector is SAM-3~\citep{sam3}
(\texttt{facebook/sam3}), used off the shelf and never finetuned, with
the same text prompts and thresholds at corpus construction and at
deployment (Appendix~\ref{app:bankquality}). The executor is the fixed
subgoal-conditioned $\pi_{0.5}$ shared with the oracle rows; only
the subgoal stream differs.
\textbf{Evaluation protocol.} Closed loop, 30 episodes per task per
serving seed, three serving seeds for the released 9B configuration, one
agent checkpoint; each episode
runs Algorithm~\ref{alg:tick} behind the harness with corpus-time
detector, prompts and window geometry reproduced verbatim. Baseline and
oracle rows are RoboMME-reported (50 episodes over three checkpoints and
three seeds); the metric is episode success under each task's
environment predicate.

\subsection{Main results}

\sys{} averages $66.0$ across the sixteen tasks
(Appendix~\ref{app:seeds} per seed), $+21.5$ over the best
prior non-privileged method (FrameSamp+Modul, $44.5$) and $78.5\%$ of the
privileged GroundSG oracle ($84.1$), and is the best non-privileged
method on nine of sixteen tasks, reaching the oracle ceiling on
PickXtimes, PatternLock and RouteStick (Table~\ref{tab:main}).

The gains concentrate where memory is a matter of \emph{identity}: on the
shuffle and reveal families, where success hinges on carrying a
role-to-instance binding across occlusion, \sys{} roughly doubles the
best baseline (VUS $75.6$ vs.\ $38.0$, VRP $65.6$ vs.\ $30.4$, VPO
$66.7$ vs.\ $36.0$), because evidence-gated move chains preserve the
binding that keyframe selection drops. It leads press-counting (BU
$84.4$, BUS $35.6$), where the press gate turns premature completions
into retries, and recovers the oracle on the route-tracing tasks (PL
$98.9$, RS $56.7$), where every prior VLM pipeline is at chance.

\begin{table}[t]
\centering
\scriptsize
\setlength{\tabcolsep}{3pt}
\resizebox{\textwidth}{!}{%
\begin{tabular}{cl|cccc|cccc|cccc|cccc|c}
\toprule
& & \multicolumn{4}{c|}{Temporal} & \multicolumn{4}{c|}{Spatial}
& \multicolumn{4}{c|}{Object} & \multicolumn{4}{c|}{Procedural} & \\
& Method & PickX & BinF & SwingX & StopC & VU & BU & VUS & BUS
& PH & VRP & VPB & VPO & MC & IP & PL & RS & Avg \\
\midrule
{\tiny\emph{e2e}} &
$\pi_{0.5}$
& 42.9 & 30.0 & 35.6 & 6.7 & 20.4 & 22.2 & 18.7 & 6.7
& 11.3 & 0.4 & 31.1 & 25.8 & 26.0 & 1.6 & 2.9 & 4.7 & 17.9 \\
& SAM2Act+
& 76.0 & 40.0 & 25.3 & 0.0 & 27.3 & 32.0 & 18.0 & 26.7
& 17.3 & 5.3 & 24.7 & 20.0 & 29.3 & 0.0 & 0.0 & 0.0 & 21.4 \\
\cmidrule{1-19}
{\tiny\emph{mem-VLA}} &
TokenDrop+Modul
& 83.6 & 34.4 & 86.0 & 5.3 & 28.2 & 29.3 & 28.4 & 21.3
& 21.3 & 22.0 & 59.6 & 36.0 & 62.0 & 7.1 & 32.4 & 51.6 & 38.0 \\
& FrameSamp+Modul
& 87.3 & 39.6 & \textbf{92.0} & \textbf{42.0} & 32.7 & 25.1 & 24.4 & 18.2
& 22.9 & 30.4 & 60.0 & 32.0 & 77.8 & \textbf{7.6} & 53.6 & \textbf{66.7} & 44.5 \\
\cmidrule{1-19}
{\tiny\emph{VLM-mem}} &
MemER
& 79.3 & 56.7 & 59.3 & 0.0 & 81.3 & 72.0 & 38.0 & 21.3
& 70.7 & 25.3 & 30.0 & 26.0 & \textbf{82.7} & 6.7 & 16.7 & 12.0 & 42.4 \\
& SimpleSG+QwenVL
& 95.3 & \textbf{77.6} & 5.1 & 0.4 & 34.2 & 19.3 & 15.3 & 9.6
& 17.1 & 25.3 & 33.3 & 25.1 & 82.0 & 3.8 & 12.7 & 7.8 & 29.0 \\
& GroundSG+QwenVL
& 92.7 & 52.0 & 7.3 & 0.0 & \textbf{88.7} & 24.0 & 30.7 & 14.0
& 15.1 & 25.3 & 54.0 & 31.8 & 71.6 & 3.3 & 6.7 & 6.0 & 32.7 \\
& GroundSG+Gemini
& 18.0 & 26.0 & 4.0 & 3.0 & 36.0 & 14.0 & 13.0 & 0.0
& 9.0 & 17.0 & 12.0 & 7.0 & 17.0 & 0.0 & 7.0 & 2.0 & 11.6 \\
\midrule
\rowcolor{oursrow}
& \sys{} (ours)
& \textbf{98.9} & 71.1 & 68.9 & 35.5 & 86.7 & \textbf{84.4} & \textbf{75.6} & \textbf{35.6}
& \textbf{71.1} & \textbf{65.6} & \textbf{75.5} & \textbf{66.7} & 62.2 & 3.3 & \textbf{98.9} & 56.7 & \textbf{66.0} \\
\rowcolor{oursrowlite}
& \sys{} (ours), \emph{0.8B agents}
& 76.7 & 54.4 & 52.2 & 40.0 & 65.6 & 76.7 & 43.3 & 25.6
& 51.1 & 68.9 & 87.8 & 77.8 & 41.1 & 0.0 & 98.9 & 58.9 & 57.4 \\
\midrule
\rowcolor{oraclerow}
& SimpleSG+Oracle (privileged)
& 99.8 & 85.8 & 100.0 & 44.7 & 33.1 & 22.0 & 15.6 & 15.6
& 44.0 & 27.8 & 31.3 & 26.0 & 87.3 & 10.0 & 95.3 & 55.1 & 49.6 \\
\rowcolor{oraclerow}
& GroundSG+Oracle (privileged)
& 100.0 & 85.8 & 100.0 & 49.7 & 98.8 & 95.0 & 99.2 & 80.2
& 83.3 & 97.3 & 100.0 & 100.0 & 87.8 & 15.6 & 97.0 & 55.6 & 84.1 \\
\bottomrule
\end{tabular}}
\caption{Closed-loop success (\%) on the sixteen memory tasks, grouped
as in RoboMME Table~3: end-to-end VLAs, memory-VLA architectures, VLM
symbolic-memory pipelines, privileged oracles (gray). Baseline and
oracle rows are RoboMME-reported; both \sys{} rows are means over three
serving seeds at 30 episodes/task/seed behind the oracle rows' own fixed
$\pi_{0.5}$ (per-seed numbers in Appendix~\ref{app:seeds}). The
\emph{0.8B agents} row swaps only the agents' base model
(Section~\ref{sec:ablation}). Bold: best non-privileged result per task,
excluding the 0.8B row.}
\vspace{-4pt}
\label{tab:main}
\end{table}

\subsection{Harness ablation: which verdict, and at which scale}
\label{sec:ablation}

We ablate along two axes (Table~\ref{tab:ablation}): one verdict at a
time at 9B, which attributes the harness's value to a kind of evidence;
and the agents' capacity with the harness on and off, which asks how much
of the result is the harness rather than the model. The contracts exist
because a fluent proposer is not a reliable reporter, so their value
should grow as reliability falls.

\textbf{Per verdict (9B).} Without \textsc{defer}'s press-verify
gate, completions take effect on the writer's schedule again and the
press-counting tasks collapse (BU $84\to57$, BUS $36\to13$); without
\textsc{reject}, chance-level direction words replace the extractor's
and both route tasks fail outright (PL $99\to3$, RS $57\to7$); without
\textsc{correct}, the binding tasks lose four to seven points, the
smallest effect, since rollout-matched training absorbs most grounding
repair.

\begin{table}[t]
\centering
\scriptsize
\setlength{\tabcolsep}{3pt}
\resizebox{\textwidth}{!}{%
\begin{tabular}{l|cccc|cccc|cccc|cccc|c}
\toprule
& \multicolumn{4}{c|}{Temporal} & \multicolumn{4}{c|}{Spatial}
& \multicolumn{4}{c|}{Object} & \multicolumn{4}{c|}{Procedural} & \\
Configuration & PickX & BinF & SwingX & StopC & VU & BU & VUS & BUS
& PH & VRP & VPB & VPO & MC & IP & PL & RS & Avg \\
\midrule
\multicolumn{18}{l}{\emph{Writer / reasoner base: Qwen3.5-9B}} \\
\sys{} (full harness)
& 98.9 & 71.1 & 68.9 & 35.5 & 86.7 & 84.4 & 75.6 & 35.6
& 71.1 & 65.6 & 75.5 & 66.7 & 62.2 & 3.3 & 98.9 & 56.7 & 66.0 \\
\quad w/o \textsc{defer} (press-verify gate)
& 95.9 & 70.1 & 67.2 & 36.8 & 86.1 & \cellcolor{defercell}56.7 & 75.4 & \cellcolor{defercell}13.1
& 71.6 & 67.9 & 73.2 & 65.6 & 64.4 & 6.5 & 97.8 & 55.3 & 62.7 \\
\quad w/o \textsc{reject} (extractor-owned claims, duplicates)
& 95.6 & 71.3 & 66.7 & \cellcolor{rejectcell}13.5 & 85.7 & 83.2 & 76.1 & 36.1
& 70.4 & 65.3 & 74.8 & 63.1 & 67.2 & 3.3 & \cellcolor{rejectcell}3.3 & \cellcolor{rejectcell}6.7 & 55.1 \\
\quad w/o \textsc{correct} (grounding rebinding)
& 98.3 & 72.3 & 66.7 & 36.1 & 85.9 & 86.2 & 76.4 & 35.1
& 70.9 & \cellcolor{correctcell}61.3 & \cellcolor{correctcell}71.4 & \cellcolor{correctcell}60.2 & 61.2 & 3.3 & 97.5 & 57.3 & 65.0 \\
\quad no harness
& 96.7 & 69.4 & 67.2 & \cellcolor{offcell}13.6 & 83.4
& \cellcolor{offcell}58.2 & 74.8 & \cellcolor{offcell}14.2
& 70.5 & \cellcolor{offcell}61.1 & \cellcolor{offcell}70.9
& \cellcolor{offcell}61.3 & 62.7 & 5.6
& \cellcolor{offcell}2.1 & \cellcolor{offcell}6.7 & 51.2 \\
\midrule
\multicolumn{18}{l}{\emph{Writer / reasoner base: Qwen3.5-0.8B}} \\
\sys{} (full harness)
& 76.7 & 54.4 & 52.2 & 40.0 & 65.6 & 76.7 & 43.3 & 25.6
& 51.1 & 68.9 & 87.8 & 77.8 & 41.1 & 0.0 & 98.9 & 58.9 & 57.4 \\
\quad no harness
& \cellcolor{offcell}70.0 & \cellcolor{offcell}46.7 & 56.7 & \cellcolor{offcell}10.0 & 73.3
& \cellcolor{offcell}53.3 & 50.0 & \cellcolor{offcell}6.7
& 50.0 & \cellcolor{offcell}26.7 & \cellcolor{offcell}13.3 & \cellcolor{offcell}23.3
& \cellcolor{offcell}30.0 & 0.0
& \cellcolor{offcell}3.3 & \cellcolor{offcell}3.3 & 32.3 \\
\bottomrule
\end{tabular}}
\caption{Harness ablation on two axes, in the layout of
Table~\ref{tab:main}. \emph{Top}: one verdict disabled at a time at 9B
(three seeds, 30 episodes/task/seed); each coloured cell loses at least
four points, in the removed verdict's colour---\textsc{defer} blue,
\textsc{reject} orange, \textsc{correct} violet. \emph{Bottom}: only the
base model is replaced; grey marks the same four-point loss on the
\emph{no harness} rows.}
\vspace{-4pt}
\label{tab:ablation}
\end{table}

\textbf{Across scale.} Scaling the agents down degrades precisely what
the harness polices: hallucination rises, spatial grounding weakens, and
identity binding across long chains of change decays. The same verdicts
therefore have more to correct, and the harness is worth $+25.1$ points at
0.8B ($32.3 \to 57.4$) against $+14.8$ at 9B ($66.0$ against $51.2$);
on the thirteen tasks whose evidence extractors are identical across
scales the small-scale figure is $+17.8$ ($34.9 \to 52.7$). A verified
small writer still exceeds the best prior non-privileged method
($57.4$ vs.\ $44.5$): the contracts, not the parameter count, carry most
of the memory competence.

\subsection{Tick-level diagnosis: auditing a VLM as a memory writer}
\label{sec:diagnosis}

\begin{figure}[t]
\centering
\includegraphics[width=0.96\linewidth]{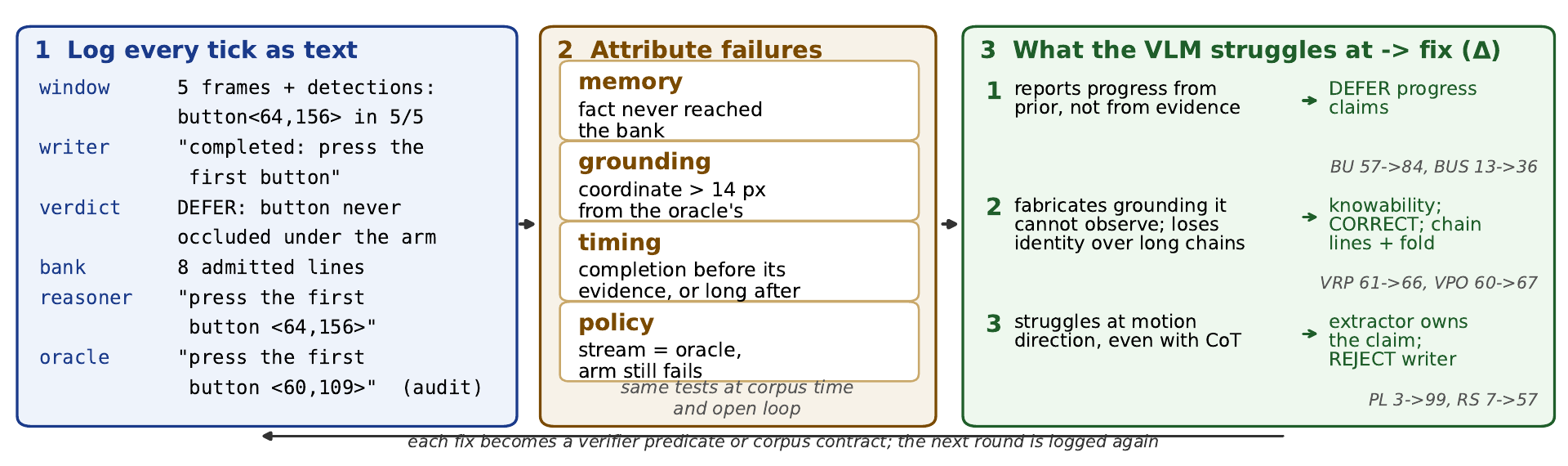}
\caption{The audit loop: every tick is logged as text (left, a real
ButtonUnmaskSwap tick whose press claim is deferred because the button is
still detected); each failed episode is attributed mechanically to one
defect class (middle); the defects become the findings on the right, each
turned into a predicate validated by its closed-loop delta.}
\vspace{-4pt}
\label{fig:audit}
\end{figure}

Because the memory is readable text, a VLM's fitness as a memory writer
can be measured rather than assumed (Figure~\ref{fig:audit}). The serve
trace records every tick (the writer's line, its verdict, the bank, the
reasoner's subgoal, the live oracle), and each failed episode is
attributed mechanically to one of four defects: a \emph{memory} defect (a
fact the bank never received), a \emph{grounding} defect (right words,
wrong object: a coordinate beyond the wrong-instance radius), a
\emph{timing} defect (a completion registered before or after its
evidence), or a \emph{policy} failure (the stream matches the oracle,
the arm still fails). The same audit runs at corpus time and open loop,
so a defect traces to data, model, or serving.

The audit yields findings about the VLM, not only numbers; each became a
predicate validated by its closed-loop delta (Table~\ref{tab:ablation};
screening rounds in Appendix~\ref{app:rounds}):
\begin{list}{}{\setlength{\leftmargin}{1.2em}\setlength{\itemsep}{0pt}\setlength{\parsep}{0pt}\setlength{\topsep}{1pt}}
\item \textbf{Finding 1: VLMs report progress from prior, not from
evidence.} A finetuned writer declares an invisible action complete
when it usually happens, not when it sees it; \textsc{defer} on press
claims raised BU $57\to84$ and BUS $13\to36$.
\item \textbf{Finding 2: VLMs fabricate grounding they cannot observe
and lose identity across long chains of change.} Grounding defects were
plausible coordinates absent from the model's inputs, and the reasoner
mis-folds long swap chains; knowability, \textsc{correct} and
deterministic per-move folds are worth four to seven points on the
binding tasks.
\item \textbf{Finding 3: VLMs struggle at judging motion direction
from frames.} A 9B VLM finetuned on 1{,}600 episodes with direction
targets, chain-of-thought supervision and six input reformulations still
read a stroke's rotation sense at chance, though the answer is one line
of arithmetic over detector output (Appendix~\ref{app:motion}). Letting
the extractor own the claim and \textsc{reject}ing the writer's raised
PL $3\to99$ and RS $7\to57$.
\end{list}
The audit of both single-VLM pipelines
(Table~\ref{tab:threeway}, Appendix~\ref{app:tickdiag}) finds binding
failing by grounding, press-counting by timing, direction by content,
MemER holding no demonstration frame.

\subsection{A second benchmark: RoboMemArena}
\label{sec:rma}

\sys{} transfers to RoboMemArena~\citep{rma}, 26 kitchen tasks in
four families, with its contracts unchanged
(Appendix~\ref{app:rma}): detector, writer and harness run as on
RoboMME behind the benchmark's own $\pi_{0.5}$ recipe; every task has
a fixed subtask sequence, so the reasoner is pinned to it and the memory
stage decides when the executor is handed the next subtask. The port
changed two geometry constants, detector prompts and five evidence
predicates, and the harness's verdicts generalise unchanged to the new
executor and evidence stream: the full pipeline reaches 38.8 task
and 59.0 subtask success on held-out layouts, above the reported
memory-VLA and keyframe methods and level with the benchmark's
end-to-end memory system (Table~\ref{tab:rma}; ablations in
Appendix~\ref{app:rma}).

\begin{table}[h]
\centering
\scriptsize
\setlength{\tabcolsep}{3pt}
\renewcommand{\arraystretch}{0.85}
\begin{tabular}{l|cccc|c}
\toprule
RoboMemArena, task / subtask success (\%) & Transferring & Occlusion & Counting & Sequence & Average \\
\midrule
$\pi_{0.5}$ (no subgoals) & 20.0 / 42.8 & 12.7 / 17.2 & 14.3 / 50.9 & 60.0 / 71.6 & 21.5 / 38.7 \\
MemoryVLA~\citep{memoryvla} & 15.0 / 37.2 & 7.3 / 13.1 & 14.3 / 55.1 & 37.5 / 65.2 & 15.0 / 35.3 \\
MemER~\citep{memer} & 20.0 / 36.1 & 16.4 / 33.2 & 27.1 / 65.1 & 65.0 / 79.1 & 27.3 / 49.1 \\
PrediMem~\citep{rma} & 22.5 / \textbf{45.2} & \textbf{27.3} / 38.4 & 45.7 / \textbf{69.3} & 72.5 / 89.5 & 38.5 / 55.2 \\
\midrule
\rowcolor{oursrow} \sys{} (ours) & \textbf{27.4} / 37.6 & 19.2 / \textbf{50.1} & \textbf{52.3} / 67.5 & \textbf{80.7} / \textbf{90.3} & \textbf{38.8} / \textbf{59.0} \\
\midrule
\rowcolor{oraclerow} ground-truth subtask feed (privileged) & 32.5 / 54.8 & 33.6 / 49.8 & 51.4 / 75.6 & 85.0 / 92.3 & 46.1 / 64.8 \\
\bottomrule
\end{tabular}
\caption{RoboMemArena by task family (the benchmark's Table~2 layout):
baseline and privileged rows as reported; \sys{} on held-out layouts,
50 episodes per task. Bold: best non-privileged result per column.}
\vspace{-4pt}
\label{tab:rma}
\end{table}

\section{Discussion and Limitations}
\label{sec:limitations}

\textbf{Simulation only.} Our contribution is an algorithm for the
memory stage, how episodic observation becomes a grounded symbolic
subgoal, and the measurements behind its contracts, not a claim about
physical manipulation, which enters only through the fixed executor and
the detector. Simulation also makes the controlled comparison possible
(a privileged ceiling, 480 held-out episodes per configuration); a real
robot would change the perception front-end and the executor, not the
contracts.

\textbf{Serving cost.} Each tick costs one detector pass and two VLM
generations; our closed-loop runs wait for the call, so the reported
success rates do not depend on its latency, and a deployment would run
the agents asynchronously at the subgoal timescale while the policy
acts at control rate. The 0.8B agents keep $57.4$ success behind the
same harness (Section~\ref{sec:ablation}), within reach of edge
hardware; event-triggered agent calls are the remaining remedy and
future work.

\clearpage
\subsection*{AI use statement}
In this work we used generative AI tools in four ways, all under the
authors' direction and with the authors responsible for every claim.
(i)~\emph{Writing}: an LLM-based coding assistant helped draft and
polish sections of the manuscript, captions, and tables from the
authors' notes and results; all text was reviewed and edited by the
authors. (ii)~\emph{Retrieval}: the same assistant was used to search
for and summarise related work; every citation was verified by the
authors against the cited source. (iii)~\emph{Research execution}:
the assistant was used to implement code under the authors'
specifications (the corpus generator, the harness's verifier
predicates, the serving stack, the evaluation lanes and the trace
analysis and figure scripts) and to help attribute closed-loop
failures from the logged traces; the design decisions (the claim
types, the verdicts and the evidence predicates) and the
interpretation of all results are the authors'. LLM-written code was
reviewed and tested by the authors, and every number we report for our
own system was produced by the released evaluation scripts on the
benchmarks' held-out episodes. (iv)~\emph{Figures}: the artwork of
Figures~\ref{fig:teaser} and~\ref{fig:system} was produced with an
image-generation model from author-written prompts and then patched
with verbatim text and frames from real episodes; all other figures
are drawn by scripts, from logged data or, for the schematics
(Figures~\ref{fig:audit} and~\ref{fig:harness-schematic}), from
author-written layouts carrying verbatim logged text. We have not used generative
AI tools for synthetic data generation (the corpus is generated by
deterministic code from the benchmark's recorded episodes) or for
proving mathematical claims; theoretical model development is not
applicable to this work. We have reviewed all AI-assisted work and
take responsibility for the final content, including text, claims and
artifacts produced with the aid of generative AI.

\subsection*{Reproducibility statement}
Our code and trained weights are available at
\url{https://github.com/1999Lyd/GAMMA}. The release contains the complete pipeline behind every
number in this paper: the SAM-3 detection cache with its per-class prompts
and thresholds; the corpus generator implementing the three train-time
contracts of Section~\ref{sec:targets} together with the verification
gates run before training (knowability audit, evidence-gate alignment,
bank statistics); the ms-swift finetuning scripts for both agents with
the exact hyperparameters of Appendix~\ref{app:config}; the writer
rollout that produces the reasoner's training banks; the agent server
implementing Algorithm~\ref{alg:harness}, in which every verdict and
every evidence extractor is a documented, individually switchable
predicate (the ablation rows of Table~\ref{tab:ablation} are produced
by these switches); the closed-loop evaluation client and lane
launchers with the serving seeds ($7$, $8$, $9$) and the $30$
episodes-per-task protocol of Section~\ref{sec:experiments}; the
per-tick trace logger that records detections, writer line, bank,
reasoner output and live oracle, which is the substrate of the tick-level
diagnosis and of every figure; and the RoboMemArena port of
Appendix~\ref{app:rma}. The released weights are the writer and reasoner
LoRA adapters and the fixed subgoal-conditioned $\pi_{0.5}$ executor
shared by all configurations; base models (Qwen3.5, SAM-3) and the
benchmark's recorded episodes are public. All machine-specific paths are
read from a single environment file, and the README walks through the
eight stages from raw episodes to the tables.

\bibliography{references}
\bibliographystyle{iclr2027_conference}

\clearpage
\appendix

\section{Harness instantiations and the limiting case}
\label{app:motion}

\textbf{Verifier instantiations.} Progress claims: a press completion is
admitted when the pressed object's detection is occluded for consecutive
frames with the arm on it; a stroke completion when the effector arrives
within a radius of the stroke's recorded endpoint; deferral otherwise,
with the plan pinned to the unverified step. Grounding claims: a cited
coordinate matching a demonstration-video location but no current
detection is rebound to the nearest class-compatible detection; a
container coordinate inconsistent with the deterministic fold of the
writer's admitted move chain is rebound to the fold. Event claims are
verified at corpus time by the same predicates (evidence gates,
knowability). Rejected claims: consecutive duplicates of an admitted
line (writer stutter), and proposer lines for claim types owned by an
authoritative evidence extractor (below), whose version supersedes the
VLM's.

\textbf{The limiting case.} Two tasks defeat any frame-based VLM by
construction. In RouteStick the
demonstration is a sequence of circling strokes around obstacles, and the
required subgoal words are \emph{clockwise/counterclockwise}; in
PatternLock the demonstration traces a path over a dot grid and the words
are eight-way directions (\emph{forward-left}, \ldots). The evidence is
continuous motion: a fast-fading ink wisp and small marker
jumps, spread over dozens of frames, with the direction determined by the
side of a bulge relative to a chord. A finetuned writer's words at
deployment were stereotyped guesses, and six
successive attempts to expose the signal to the VLM---input
reformulations and chain-of-thought supervision on the direction
judgment---measured at chance.

\begin{figure}[tb]
\centering
\includegraphics[width=0.98\linewidth]{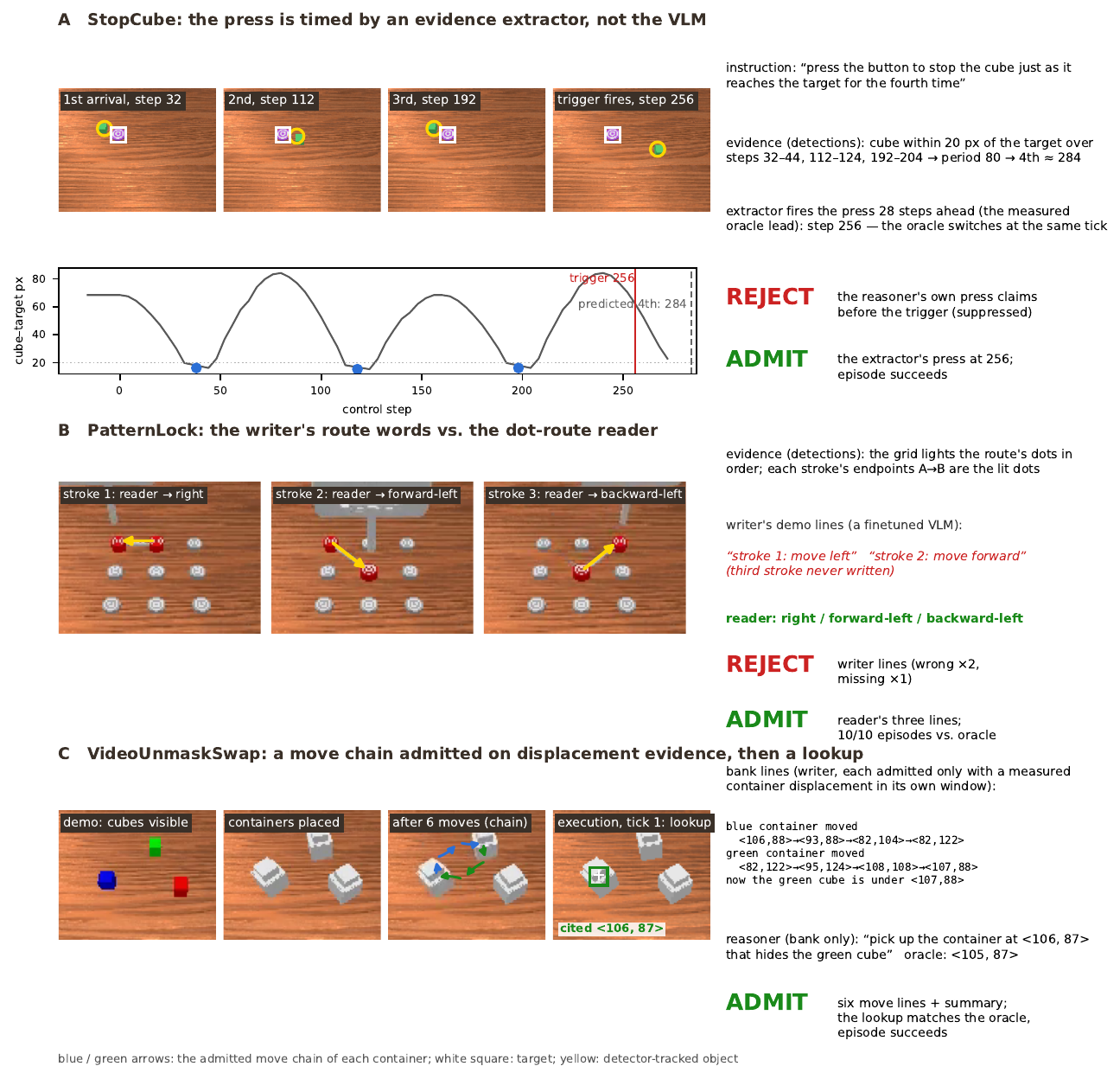}
\caption{Three more harness cases from real episodes (9B agents,
serving seed 7), in the grammar of Figure~\ref{fig:harness-cases}.
\textbf{A}, StopCube: the press is an extractor-owned claim---the
cube's arrivals at the target are counted from detections, the next
one is predicted from the observed period, and the press fires at the
measured oracle lead; the reasoner's own press claims are rejected
until then. \textbf{B}, PatternLock: the dot-route reader takes the
lit-dot sequence as stroke endpoints; the writer's words are wrong on
both strokes it wrote and it never wrote the third, so its lines are
rejected and the reader's admitted. \textbf{C}, VideoUnmaskSwap: the
shuffle is written as a chain of six move lines, each admitted only
with a measured container displacement in its own window, plus the
summary; the reasoner's lookup at execution matches the oracle's
coordinate to within one pixel.}
\label{fig:harness-more}
\end{figure}

We therefore treat motion grounding exactly as we treat spatial
grounding---as a tool. Both demonstrations turn out to be
mechanically decodable: strokes arrive in fixed 50-frame segments; a
rendered marker stands at each stroke's start pad and teleports at
boundaries; the ink's fresh pixels sample the stroke's arc; and the
pattern grid lights its route dots in order. Small deterministic readers
recover every stroke's endpoints and orientation
(sign of the median cross-product of ink samples about the directed
chord), scoring $33/33$ strokes on RouteStick and $10/10$ episodes on
PatternLock against the development annotations
(Algorithm~\ref{alg:readers}; agreement with the live oracle on the
evaluation demonstrations in Table~\ref{tab:readers}). At deployment the
writer's demo-record lines for these two tasks are emitted by the readers
in the same numbered format the training corpus uses, and the reasoner's
stroke-by-stroke progress is advanced deterministically by arrival
detection at the known next endpoint. The division of labour is
unchanged in spirit: the VLM writes and reasons over symbols; whatever
perception at which the VLM measurably fails is delegated to a tool.
Figure~\ref{fig:harness-more} shows three further cases from real
episodes: the extractor-owned press timing of StopCube, the
PatternLock reader, and an admitted move chain.

\begin{figure}[b]
\centering
\includegraphics[width=0.62\linewidth]{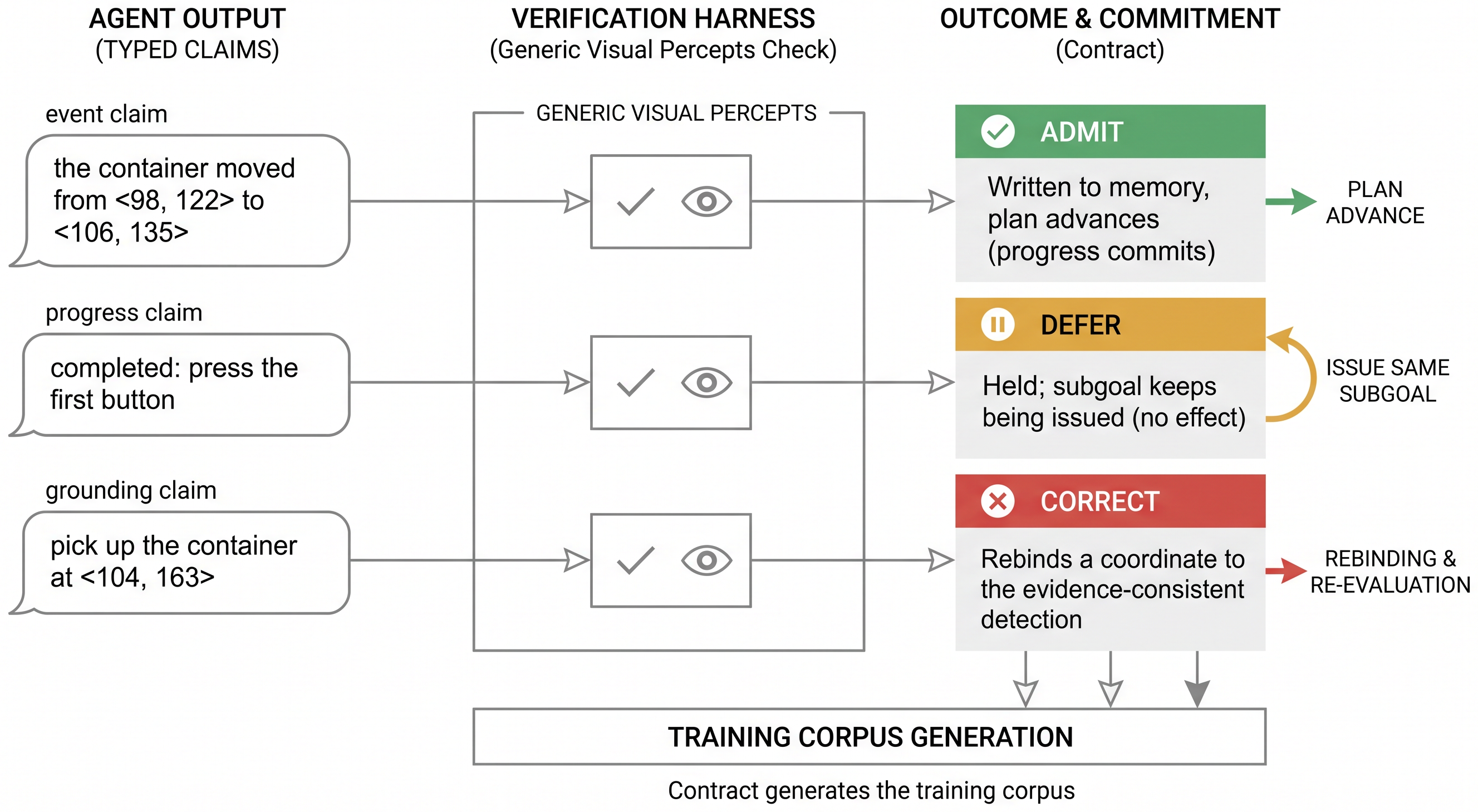}
\caption{The propose--verify harness (companion to
Algorithm~\ref{alg:harness}), one example verdict per claim type:
\textsc{admit} commits the claim and advances the plan; \textsc{defer}
holds it with no effect---the same subgoal keeps being issued until
its evidence appears; \textsc{correct} rebinds a mis-grounded
coordinate to the evidence-consistent detection (\textsc{reject}
omitted for clarity). The same contract generates the training corpus.}
\label{fig:harness-schematic}
\end{figure}

\textbf{On generality.} The readers encode benchmark regularities (fixed
stroke length, marker semantics). We regard this as the current boundary of
VLMs---the readers replace percepts that six learned
formulations measured at chance---but transferring the system to a new
environment requires either re-deriving such readers or a VLM that can
genuinely read motion; the second is the open research problem.

\begin{algorithm}[tb]
\caption{Deterministic demonstration readers (RouteStick, PatternLock)}
\label{alg:readers}
\scriptsize
\begin{algorithmic}[1]
\Require demonstration frames $o_{0:T}$ (RGB); stroke length $S=50$
\Statex \textbf{RouteStick} (stroke sense and target side)
\State $K \gets \mathrm{round}(T/S)$; for each stroke $k$, $A_k \gets$ median
  position of the red marker over frames $[kS, kS{+}40)$ \Comment{marker stands on the start pad}
\State chord direction $\hat v \gets$ normalised $A_1{-}A_0$, spacing $\gets$ median $\lVert A_{k+1}{-}A_k\rVert$
\For{$k = 0 \ldots K{-}1$}
  \State $P_k \gets$ ink pixels: new white pixels replacing wood between consecutive frames
    of stroke $k$, in components $\le 150$\,px, within $55$\,px of $A_k$
  \State $B_k \gets A_{k+1}$ if the next marker exists, else $A_k + \hat v\cdot\text{spacing}\cdot\mathrm{sign}(\mathrm{median}\ \text{proj})$
  \State $m_k \gets \mathrm{median}_{(x,y)\in P_k}\ \big[(B_k{-}A_k)\times((x,y){-}A_k)\big]$ \Comment{cross product, image coords}
  \State sense $\gets$ clockwise if $m_k<0$ else counterclockwise;\; side $\gets$ left if $(B_k{-}A_k)_x>0$ else right
  \State emit \emph{demo showed: stroke $k{+}1$: move to the nearest \{side\} target by circling around the stick \{sense\}}
\EndFor
\Statex \textbf{PatternLock} (eight-way direction per stroke)
\State route $\gets$ ordered list of red dots as they first light up (stride 2, new dot if $>8$\,px from all previous)
\For{consecutive dots $(a_k, a_{k+1})$}
  \State $\Delta \gets a_{k+1}-a_k$; fore $\gets$ forward/backward if $|\Delta_{\text{row}}|\ge 9$; lat $\gets$ left/right if $|\Delta_{\text{col}}|\ge 10$
  \State emit \emph{demo showed: stroke $k{+}1$: \{fore\}-\{lat\}} (whichever components are non-empty)
\EndFor
\Statex At execution, stroke $k$ is advanced when the effector arrives within the
  hit radius of $B_k$ (RouteStick) or dot $a_{k+1}$ lights (PatternLock).
\end{algorithmic}
\end{algorithm}

\textbf{Validation on the evaluation demonstrations.} Because the
demonstration of an evaluation episode is fixed by its index, the three
serving seeds replay the same 30 demonstrations per task; the reader's
record can therefore be scored against the live oracle's stroke words
on every stroke the policy reached. Table~\ref{tab:readers} gives the
seed-7 agreement (seeds 8 and 9 reproduce it on the same demonstrations).
The RouteStick residue is concentrated in seven demonstrations
(episodes 3, 10, 15, 19, 22, 26 and 27), in each of which the reader
mis-reads one or more strokes; RouteStick success ($56.7$) nevertheless
matches the privileged oracle's ($55.6$), so the policy, not the
record, is the ceiling there.

\begin{table}[tb]
\centering
\footnotesize
\setlength{\tabcolsep}{5pt}
\begin{tabular}{lrrrrr}
\toprule
Task (seed 7, 30 episodes) & strokes reached & sense correct & side correct & both & episodes fully correct \\
\midrule
RouteStick  & 79 & 69 & 73 & 66 & 23/30 \\
PatternLock & 85 & --  & --  & 80 & 27/30 \\
\bottomrule
\end{tabular}
\caption{Reader record vs.\ the live oracle on the held-out evaluation
demonstrations: strokes are counted once the oracle issues them; for
PatternLock a stroke is correct when its eight-way word matches. On the
development annotations the readers score $33/33$ strokes and $10/10$
episodes respectively.}
\label{tab:readers}
\end{table}

\textbf{Serve-side guard inventory.} Table~\ref{tab:guards} lists every
serve-time guard of the released server, the claim type it owns, the
evidence it consults and the verdict it issues. Trigger counts are
given where the seed-7 run logged them; the remaining guards were not
instrumented in that run (they are in the released server).

\begin{table}[tb]
\centering
\scriptsize
\setlength{\tabcolsep}{4pt}
\begin{tabular}{>{\raggedright\arraybackslash}p{2.8cm} >{\raggedright\arraybackslash}p{1.9cm} >{\raggedright\arraybackslash}p{4.4cm} >{\raggedright\arraybackslash}p{1.4cm} >{\raggedright\arraybackslash}p{1.8cm}}
\toprule
Guard & Claim type & Evidence & Verdict & Triggers (seed 7) \\
\midrule
Press-verify gate & progress (press) & pressed object still detected in the window & \textsc{defer} & not logged \\
Press latch & progress (press) & object occluded under the arm for consecutive frames & \textsc{admit} & not logged \\
Arrival-timed press (StopCube) & progress (press) & counted arrivals + predicted period; writer's press rejected until then & \textsc{reject} / extractor & not logged \\
Stale-coordinate re-snap & grounding & cited coordinate matches a demo location but no current detection; nearest class-compatible detection & \textsc{correct} & not logged \\
Container-coordinate authority & grounding & deterministic fold of the admitted move chain disagrees with the cited container & \textsc{correct} & not logged \\
Direction words (RouteStick) & demo record & reader's stroke sense/side supersede the writer's & \textsc{reject} writer & 1{,}640 words / 30 eps \\
Direction words (PatternLock) & demo record & reader's eight-way word supersedes the writer's & \textsc{reject} writer & 1{,}466 words / 30 eps \\
Stroke advance & progress (stroke) & effector within the hit radius of the recorded endpoint & \textsc{admit} & not logged \\
Duplicate line & event & consecutive repeat of an admitted line & \textsc{reject} & not logged \\
\bottomrule
\end{tabular}
\caption{Serve-side guards of the released server (9B configuration).
Word counts are the number of writer/reasoner direction words
overridden by the reader over the 30 seed-7 episodes of each task
(every tick re-issues the stroke, so one stroke accrues many
overrides).}
\label{tab:guards}
\end{table}

\section{Target construction for the writer and the reasoner}
\label{app:targets}

The training targets are derived from recorded episodes by a
deterministic generator that knows nothing about individual episodes:
it takes a role vocabulary per task (which object classes matter and
what role each can play) and the general event schema below. Adding a
task means naming its roles; nothing is annotated by hand.

\textbf{Line schema.} Every writer target is either \textsc{none} or one
or more lines of the form \emph{[event] kind: sentence [sam] evidence},
where \emph{kind} is one of four:
\begin{list}{}{\setlength{\leftmargin}{1.2em}\setlength{\itemsep}{1pt}}
\item \emph{scene}: the objects present and their coordinates, written
at tick 0 and again when the execution phase begins;
\item \emph{demo showed}: one line per demonstrated step (a pick, a drop
onto a target, a stroke with its direction words), in the order shown;
\item \emph{observed}: a change the window makes visible---an object
appearing, a container moving from one coordinate to another, an object
reaching or leaving a location, and, after a sequence of moves, a
summary of the resulting state (which cube is under which container);
\item \emph{completed}: the sub-task the policy has just finished,
phrased as the oracle's subgoal string.
\end{list}
The sentence names objects by role (\emph{the red cube's container},
\emph{the correct cube}) rather than by appearance, and the
\emph{[sam]} suffix lists the detections the sentence rests on, so a
line is auditable against its own window.

\textbf{Sources.} Three sources supply candidate events. The
benchmark's privileged subgoal stream gives the completion boundaries
(a \emph{completed} line where the stream switches) and the
demonstrated steps. Offline trackers over the detector's per-frame
output supply the events the benchmark never annotates: each tracker
is a deterministic function of the detections (a container's centroid
track yields \emph{moved} lines; a cube's disappearance under a
container yields the covering summary; a new detection of a class
yields an \emph{appeared} line). The detector alone supplies the scene
lines. Where the oracle and a tracker describe the same event the oracle
is used to verify and correct the tracker, never to replace it.

\textbf{Placement and evidence.} A candidate event becomes a target only
at the window whose frames contain its evidence, under the predicate of
its kind: an appearance requires the object to be newly detected in the
window, a move requires the detected endpoints to differ by at least
the displacement threshold, a completion requires the acting motion to
be visible. Annotation timing therefore never matters: an event
annotated one window early or late is simply attached to the window
that shows it, and a candidate with no evidencing window is dropped
rather than placed by schedule. Windows with no evidenced event have the
target \textsc{none}, which is the majority of ticks
(Appendix~\ref{app:bankquality}).

\textbf{Grounding and knowability.} Every coordinate in a target is a
detector reading from the evidencing window or a coordinate already
present in the bank lines the writer is shown; the generator checks this
for every target (the knowability audit) and demotes a line whose
coordinate cannot be found to its coordinate-free form. The writer is
thus never trained to produce a coordinate it could not have read.

\textbf{Record layout.} A writer record pairs the target with exactly the
inputs the server provides at deployment: the instruction, the phase,
the reasoner's previous target (the planner's expected next subgoal),
the last eight bank lines (ground-truth lines in stage one), and the
window's frames with their detection strings in the same geometry
(Appendix~\ref{app:config}). A reasoner record pairs the instruction
and the full bank with the oracle string in progress at the next tick,
its coordinates snapped to the bank; during the demonstration the target
is the hold sentence, when the bank cannot yet support a grounded
subgoal it is \emph{hold and observe the scene}, and after the last
annotated step it is the terminal sentence. In stage two the reasoner's
banks are those written by the trained writer rolled out over the
training episodes, so its inputs carry the writer's actual omissions
and phrasings.

\section{Development evidence: closed-loop rounds}
\label{app:rounds}

The system was iterated through eight closed-loop evaluation rounds under a
fixed discipline: one change set per round, each validated by a per-task
delta on a 10-episode-per-task screening protocol (numbers in this
appendix are screening measurements; final-table numbers use the
30-episode protocol). The screening evidence behind the findings of
Section~\ref{sec:diagnosis}, and the engineering lessons referenced
there:

\textbf{Completion registration belongs to the writer.} Moving
sub-task-completion registration into the reasoner collapsed the counting
tasks: the reasoner inherited a content-clock bias and declared
completions on schedule rather than on evidence. Reverting registration
to the writer (with a one-tick delay for press events whose effect
renders invisibly) restored them.

\textbf{Object permanence as a chain.} Recording shuffles as one final
summary failed under multi-swap episodes; recording every move with real
detected endpoints, a displacement evidence gate, and the summary
after the moves took the hardest shuffle task from $4/10$ to
$10/10$, held across all subsequent rounds. A deterministic fold of the
writer's own chain further serves as a coordinate authority when the
reasoner mis-folds long chains.

\textbf{Train/serve geometry.} Training demo windows at 8 steps while
serving at 16 silently broke both video-placement tasks
(VideoPlaceButton $8/10\!\to\!2/5$, VideoPlaceOrder $5/10\!\to\!0/3$
before round v5 was stopped) and was invisible in open-loop metrics;
restoring the verbatim training-side pairing recovered them
(Table~\ref{tab:rounds}).

\textbf{Anticipatory targets for invisible effects.} A press that
registers with zero pixel change cannot be completed from evidence;
period-predicting the $N$-th object arrival and firing one tick early
took the counting-press task from $2/10$ to $5/10$. This predictor is
the one intervention that anticipates rather than verifies---a counting
heuristic outside the evidence contract, and we flag it as such.

\textbf{Delegating unlearnable percepts} moved RouteStick from $0/10$ to
$4/10$ with the deterministic demo record alone, before the arrival-based
stroke advance (Appendix~\ref{app:motion}).

\textbf{No pixels for the reasoner.} Giving the reasoner the current
frame alongside the bank improved no task in screening: progress
judgment already belongs to the writer and the harness, and fresh
coordinates reach the reasoner through admitted lines, so the frame
adds context without adding evidence.

\begin{table}[tb]
\centering
\scriptsize
\setlength{\tabcolsep}{4pt}
\begin{tabular}{lcccccccc}
\toprule
Task & v1 & v2 & v3 & v4 & v5 & v6 & v7 & v8 \\
\midrule
PickXtimes       & 9 & 6 & 8 & 8 & 9 & 10 & 9 & 10 \\
BinFill          & 9 & 2 & 7 & 6 & 8 & 8 & 6 & 7 \\
SwingXtimes      & 7 & 3 & 6 & 1 & 6 & 5 & 8 & 8 \\
StopCube         & 3 & 3 & 1 & 1 & 2 & 5 & 5 & 5 \\
VideoUnmask      & 10 & 8 & 10 & 10 & 10 & 10 & 10 & 8 \\
ButtonUnmask     & 5 & 8 & 6 & 7 & 9 & 7 & 7 & 9 \\
VideoUnmaskSwap  & 3 & 3 & 4 & 10 & 9 & 6 & 8 & 8 \\
ButtonUnmaskSwap & 0 & 1 & 1 & 0 & 0 & 1 & 3 & 5 \\
PickHighlight    & 5 & 7 & 8 & 8 & 8 & 8 & 8 & 8 \\
VideoRepick      & 2 & 1 & 2 & 4 & 6 & 9 & 8 & 8 \\
VideoPlaceButton & 9 & 8 & 9 & 8 & 2/5$^\dagger$ & 8 & 7 & 7 \\
VideoPlaceOrder  & 6 & 6 & 9 & 5 & 0/3$^\dagger$ & 7 & 5 & 6 \\
MoveCube         & 6 & 4 & 9 & 7 & 5 & 7 & 8 & 7 \\
InsertPeg        & 0 & 0 & 1 & 1 & 0 & 1 & 0 & 2 \\
PatternLock      & 1 & 0 & 1 & 0 & 1 & 1 & 0 & 10 \\
RouteStick       & 0 & 0 & 0 & 1 & 1 & 0 & 0 & 5 \\
\bottomrule
\end{tabular}
\caption{Closed-loop screening by round: successes out of 10 episodes
per task (seed 7). v1 introduced the two-agent system with the SAM-3
detector; v2 moved completion registration to the reasoner (reverted
in v3); v4 recorded shuffles as move chains; v5 exposed the
train/serve geometry mismatch ($^\dagger$ stopped after 5 and 3
episodes); v6 added the arrival predictor and the fold authority; v7
is the released corpus and checkpoints; v8 added the deterministic
demonstration readers and the press latch. The v8 column is the first
10 episodes of the final seed-7 run (Table~\ref{tab:seeds}).}
\label{tab:rounds}
\end{table}

\section{Tick-level diagnosis: protocol, three-way statistics and case studies}
\label{app:tickdiag}

\begin{figure}[tb]
\centering
\includegraphics[width=\linewidth]{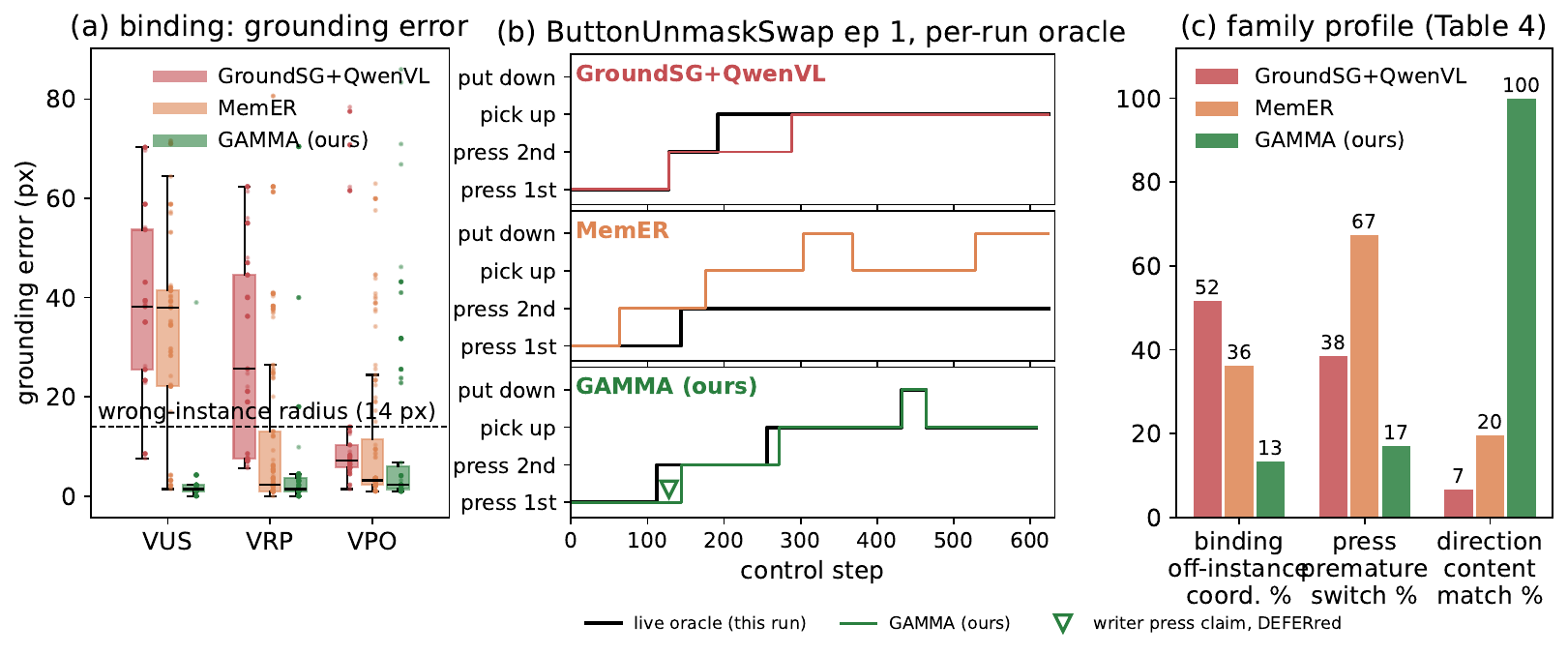}
\caption{Three-way tick-level diagnosis (10 episodes/task, seed 7).
\textbf{(a)}~Grounding error on content-matched ticks for the three
binding tasks: both single-VLM pipelines put most cited coordinates
beyond the $14$\,px wrong-instance radius (dashed)---the right words
about the wrong object---while \sys{}'s coordinates are detector
readings carried by admitted bank lines and land within a few pixels.
\textbf{(b)}~ButtonUnmaskSwap episode 1, each system against the live
oracle of its own run (Table~\ref{tab:case-bus}): MemER declares the
second press five ticks before its oracle registers the first and
never returns; \sys{}'s writer claims the first press at step 128,
the claim is deferred while the button is still detected and admitted
one tick later, and the stream follows the oracle thereafter.
\textbf{(c)}~The family profile of Table~\ref{tab:threeway}: binding
fails by grounding, press-counting by timing, direction by content.}
\label{fig:tickdiag}
\end{figure}

\textbf{Protocol.} We ran RoboMME's released GroundSG+QwenVL
(Qwen3-VL-4B with the official grounded-subgoal adapter, driven by its
own evaluation code behind the same fixed policy) and the released
MemER pipeline for 10 episodes on each of their memory tasks, and
logged \sys{}'s own serve traces on the same episode indices (seed 7),
recording at every replan tick the system's prediction and the live
oracle subgoal. The live oracle is evaluated inside each run, so every
system is scored against the oracle stream of its own rollout; the
episode index fixes the scene, the demonstration and the shuffle, so
the three runs of one index see the same evidence. \emph{Content} is
exact string match after removing coordinates (all three systems target
the same oracle string recipe). \emph{Grounding error} is the pixel distance between cited
and oracle coordinates on content-matched ticks; $14$\,px, the
detector-audit hit radius, serves as the wrong-instance threshold.
\emph{Adoption} asks whether, and how many ticks late, each oracle
transition's text is ever matched. A predicted switch is
\emph{premature} when its text has not yet appeared in the oracle
stream. Table~\ref{tab:tickdiag} gives GroundSG+QwenVL's per-task
numbers and Figure~\ref{fig:tickdiag} the three-way distributions; the
family profile is
disjoint---binding fails by grounding alone, direction by content,
press-counting by timing---so each failure meets a different harness
verdict.

\begin{table}[tb]
\centering
\footnotesize
\setlength{\tabcolsep}{4pt}
\begin{tabular}{lrrrrrrrr}
\toprule
Task & Ticks & Content\% & Gnd med px & ${>}14$px\% & Trans.\ &
Adopted\% & Lag & Premature\% \\
\midrule
ButtonUnmask & 237 & 60.3 & 5.7 & 37.8 & 9 & 100.0 & 2 & 10.0 \\
ButtonUnmaskSwap & 680 & 37.4 & 7.8 & 41.3 & 17 & 100.0 & 1 & 48.3 \\
StopCube & 154 & 72.1 & 5.0 & 0.0 & 20 & 50.0 & 0 & 0.0 \\
VideoUnmaskSwap & 77 & 100.0 & 38.2 & 90.9 & 0 & -- & -- & 0.0 \\
VideoRepick & 114 & 93.9 & 25.6 & 69.5 & 6 & 100.0 & 1 & 0.0 \\
VideoPlaceOrder & 124 & 98.4 & 7.1 & 13.1 & 10 & 100.0 & 0 & 0.0 \\
PatternLock & 133 & 1.5 & -- & -- & 1 & 0.0 & -- & 0.0 \\
RouteStick & 66 & 16.7 & -- & -- & 3 & 33.3 & 0 & 66.7 \\
\bottomrule
\end{tabular}
\caption{Tick-level diagnosis of GroundSG+QwenVL (10 episodes/task,
one seed). Binding tasks: near-perfect content, off-instance
grounding (VUS: $91\%$ of coordinates beyond the radius---and its
episodes never reach the first oracle transition). Direction tasks:
content collapse. Press-counting: missed (StopCube adopts half the
transitions) and premature switches (BUS: $48\%$ premature despite
the implementation's hard-coded hold periods).}
\label{tab:tickdiag}
\end{table}

Table~\ref{tab:threeway-tasks} extends the diagnosis to MemER and to
\sys{}'s own logged stream (serve traces carry the reasoner's output and
the live oracle per tick), giving the per-task numbers behind
Table~\ref{tab:threeway}. Over the six binding and press tasks, MemER's
keyframe memory holds $1$--$9$ execution frames per episode ($1.0$ on VideoUnmaskSwap, $2.1$--$2.5$ on
the other binding tasks, $8$--$9$ on the press tasks), covers $41$ of
$59$ oracle transitions within $16$ steps, and never contains a
demonstration frame: in its released implementation the demonstration is
attached to every query as a raw video clip, outside the keyframe
memory, so the decisive shuffle is never distilled into a retrievable
fact.

\begin{table}[tb]
\centering
\scriptsize
\setlength{\tabcolsep}{4pt}
\begin{tabular}{llrrr}
\toprule
Family (tasks) & Tick-level metric & GroundSG+QwenVL & MemER & \sys{} \\
\midrule
Binding (VUS, VRP, VPO) & coordinates beyond wrong-instance radius (\%) $\downarrow$ & 51.7 & 36.2 & \textbf{13.5} \\
Press-counting (BU, BUS) & premature subgoal switches (\%) $\downarrow$ & 38.5 & 67.4 & \textbf{17.0} \\
Direction (PL, RS) & subgoal text matches oracle (\%) $\uparrow$ & 6.5 & 19.5 & \textbf{100.0} \\
\bottomrule
\end{tabular}
\caption{Tick-level diagnosis of the single-VLM pipelines and \sys{} on
their shared tasks (10 episodes/task, seed 7): each fails at the stage a
harness verdict owns.}
\vspace{-4pt}
\label{tab:threeway}
\end{table}

\begin{table}[tb]
\centering
\scriptsize
\setlength{\tabcolsep}{3.5pt}
\begin{tabular}{l|rrr|rrr|rrr}
\toprule
& \multicolumn{3}{c|}{GroundSG+QwenVL} & \multicolumn{3}{c|}{MemER} & \multicolumn{3}{c}{\sys{}} \\
Task & content & ${>}14$px & prem.\ & content & ${>}14$px & prem.\ & content & ${>}14$px & prem.\ \\
\midrule
ButtonUnmask & 60.3 & 37.8 & 10.0 & 45.6 & 7.7 & 64.9 & 70.6 & 0.0 & 28.6 \\
ButtonUnmaskSwap & 37.4 & 41.3 & 48.3 & 50.5 & 35.9 & 69.2 & 57.7 & 2.5 & 12.8 \\
StopCube & 72.1 & 0.0 & 0.0 & -- & -- & -- & 89.1 & 0.0 & 23.8 \\
VideoUnmaskSwap & 100.0 & 90.9 & 0.0 & 100.0 & 81.2 & 0.0 & 82.6 & 1.0 & 40.0 \\
VideoRepick & 93.9 & 69.5 & 0.0 & 74.4 & 24.4 & 5.3 & 83.1 & 11.9 & 8.3 \\
VideoPlaceOrder & 98.4 & 13.1 & 0.0 & 92.3 & 25.0 & 0.0 & 64.4 & 24.6 & 5.3 \\
PickHighlight & -- & -- & -- & 33.1 & 8.6 & 76.5 & 49.8 & 4.8 & 31.8 \\
PatternLock & 1.5 & -- & 0.0 & 18.8 & -- & 75.0 & 100.0 & -- & 0.0 \\
RouteStick & 16.7 & -- & 66.7 & 20.0 & -- & 77.8 & 100.0 & -- & 0.0 \\
\bottomrule
\end{tabular}
\caption{Per-task tick-level diagnosis of the three systems on the
tasks each was run on (10 episodes/task, seed 7): per-tick subgoal
\emph{content} match with the live oracle (\%), share of content-matched
coordinates beyond the $14$\,px wrong-instance radius (\%), and share of
predicted switches that are \emph{premature} (\%). Content is a strict
per-tick string match: \sys{}'s deferred switches lower it on the
binding tasks while it adopts $91$--$100\%$ of oracle transitions with
zero median lag. ``--'': not run or no coordinate in the subgoal.}
\label{tab:threeway-tasks}
\end{table}

\subsection*{Case studies}
\label{app:cases}

Tables~\ref{tab:case-bus}--\ref{tab:case-rs} print the logged tick
streams of one episode per family, seen by all three systems (same
episode index, hence the same scene and demonstration; the live oracle
is that of each system's own run). Rows are shown where the oracle or
the prediction changes, or where \sys{}'s writer makes a progress
claim; observation lines admitted to the bank are summarised by count.

\textbf{Press-counting (Table~\ref{tab:case-bus}).} MemER switches to
\emph{press the second button} at step 64 while its oracle holds the
first press until step 144: a completion declared on schedule, not on
evidence; it then advances to the pick-up at step 176 while the oracle
still waits for the second press, and never returns. GroundSG+QwenVL
matches its oracle's timing on the first press but cites
$\langle 94,118\rangle$ for the container hiding the red cube---the
container's position \emph{before} the shuffle---37\,px from the
oracle's $\langle 109,84\rangle$; MemER cites the same pre-shuffle
container ($\langle 97,122\rangle$).
\sys{}'s writer proposes \emph{completed: press the first button} at
step 128; the button is still detected in that window, so the claim is
deferred and the reasoner keeps issuing the first press; the claim
lands one tick later and the stream follows the oracle from there. Its
pick-up cites $\langle 110,85\rangle$, $1.4$\,px from the oracle,
because the bank carries the shuffle as admitted move lines and the
resulting state.

\textbf{Binding (Table~\ref{tab:case-vrp}).} The demonstration picks
the cube at $\langle 92,128\rangle$ and drops it; the correct cube then
moves twice before execution begins. Both single-VLM pipelines cite a
coordinate near the cube's demonstration-time position
($\langle 82,123\rangle$ and $\langle 89,124\rangle$, $22$--$24$\,px
from the oracle's $\langle 83,101\rangle$)---a different one of the
three identical blue cubes---and both episodes fail within eight ticks.
\sys{}'s bank holds the demo line and the two admitted move lines, the
reasoner cites $\langle 84,102\rangle$ ($1.4$\,px), and every pick and
put cycle is advanced by an admitted completion claim.

\textbf{Direction (Table~\ref{tab:case-rs}).} The demonstration's three
strokes are clockwise, counterclockwise, counterclockwise.
GroundSG+QwenVL emits \emph{counterclockwise} for the first stroke and
fails on it; MemER gets the first stroke, then keeps \emph{clockwise}
for both later strokes although it updates the target side
correctly---the sense word is a stereotype, the side is read. \sys{}'s
demo record, written by the deterministic reader
(Appendix~\ref{app:motion}), holds all three senses; stroke progress
is advanced by arrival detection (the admitted \emph{completed} lines),
and the stream matches the oracle at every tick.

\begin{table}[tb]
\centering
\scriptsize
\setlength{\tabcolsep}{3pt}
\begin{tabular}{r >{\raggedright\arraybackslash}p{4.0cm} >{\raggedright\arraybackslash}p{4.0cm} >{\raggedright\arraybackslash}p{4.5cm}}
\toprule
step & live oracle (this run) & predicted subgoal & writer claim $\to$ verdict \\
\midrule
\multicolumn{4}{l}{\textbf{GroundSG+QwenVL}} \\
0 & press the first button at $\langle64,154\rangle$ & press the first button at $\langle62,146\rangle$ &  \\
128 & press the second button at $\langle61,103\rangle$ & press the second button at $\langle58,99\rangle$ &  \\
192 & pick up the container at $\langle109,84\rangle$ that hides the red cube & press the second button at $\langle58,99\rangle$ &  \\
288 & pick up the container at $\langle109,84\rangle$ that hides the red cube & pick up the container at $\langle94,118\rangle$ that hides the red cube &  \\
\multicolumn{4}{l}{\emph{outcome:} failure (episode ends at step 1312)} \\
\midrule
\multicolumn{4}{l}{\textbf{MemER}} \\
0 & press the first button at $\langle64,154\rangle$ & press the first button at $\langle62,153\rangle$ &  \\
64 & press the first button at $\langle64,154\rangle$ & press the second button at $\langle60,102\rangle$ &  \\
144 & press the second button at $\langle64,103\rangle$ & press the second button at $\langle60,102\rangle$ &  \\
176 & press the second button at $\langle64,103\rangle$ & pick up the container at $\langle97,122\rangle$ that hides the red cube &  \\
304 & press the second button at $\langle64,103\rangle$ & put down the container &  \\
\multicolumn{4}{l}{\emph{outcome:} failure (episode ends at step 1312)} \\
\midrule
\multicolumn{4}{l}{\textbf{\sys{}}} \\
0 & press the first button at $\langle64,154\rangle$ & press the first button at $\langle66,155\rangle$ &  \\
112 & press the second button at $\langle61,103\rangle$ & press the first button at $\langle66,155\rangle$ & 3 move lines $\to$ \textsc{admit} \\
128 & press the second button at $\langle61,103\rangle$ & press the first button at $\langle65,155\rangle$ & completed: press the first button at $\langle66,155\rangle$ $\to$ \textsc{defer} \\
144 & press the second button at $\langle61,103\rangle$ & press the second button at $\langle62,104\rangle$ & completed: press the first button at $\langle66,155\rangle$ $\to$ \textsc{admit} (held claim lands) \\
256 & pick up the container at $\langle109,84\rangle$ that hides the red cube & press the second button at $\langle63,104\rangle$ &  \\
272 & pick up the container at $\langle109,84\rangle$ that hides the red cube & pick up the container at $\langle110,85\rangle$ that hides the red cube & completed: press the second button at $\langle63,104\rangle$ $\to$ \textsc{admit} \\
\multicolumn{4}{l}{\emph{outcome:} success (episode ends at step 624)} \\
\bottomrule
\end{tabular}

\caption{Case study, press-counting: ButtonUnmaskSwap episode 1 (seed
7) as seen by the three systems. Live oracle and prediction per tick;
for \sys{}, the writer's claims and their verdicts.}
\label{tab:case-bus}
\end{table}

\begin{table}[tb]
\centering
\scriptsize
\setlength{\tabcolsep}{3pt}
\begin{tabular}{r >{\raggedright\arraybackslash}p{4.0cm} >{\raggedright\arraybackslash}p{4.0cm} >{\raggedright\arraybackslash}p{4.5cm}}
\toprule
step & live oracle (this run) & predicted subgoal & writer claim $\to$ verdict \\
\midrule
\multicolumn{4}{l}{\textbf{GroundSG+QwenVL}} \\
0 & pick up the correct cube at $\langle83,101\rangle$ for the first time & pick up the correct cube at $\langle82,123\rangle$ for the first time &  \\
\multicolumn{4}{l}{\emph{outcome:} failure (episode ends at step 128)} \\
\midrule
\multicolumn{4}{l}{\textbf{MemER}} \\
0 & pick up the correct cube at $\langle83,101\rangle$ for the first time & pick up the correct cube at $\langle89,124\rangle$ for the first time &  \\
\multicolumn{4}{l}{\emph{outcome:} failure (episode ends at step 112)} \\
\midrule
\multicolumn{4}{l}{\textbf{\sys{}}} \\
demo & \multicolumn{3}{p{12.7cm}}{bank after the demonstration replay: demo showed: pick up the cube at $\langle92,128\rangle$; demo showed: drop the cube on the table; observed: the correct cube moved from $\langle92,126\rangle$ to $\langle101,110\rangle$; observed: the correct cube moved from $\langle101,110\rangle$ to $\langle89,102\rangle$} \\
0 & pick up the correct cube at $\langle83,101\rangle$ for the first time & pick up the correct cube at $\langle84,102\rangle$ for the first time &  \\
128 & put it down & put it down & completed: pick up the correct cube at $\langle84,101\rangle$ for the first time $\to$ \textsc{admit} \\
160 & pick up the correct cube at $\langle84,101\rangle$ for the second time & put it down &  \\
176 & pick up the correct cube at $\langle84,101\rangle$ for the second time & pick up the correct cube at $\langle84,101\rangle$ for the second time & completed: put it down at $\langle90,127\rangle$ $\to$ \textsc{admit} \\
\multicolumn{4}{l}{\emph{outcome:} success (episode ends at step 560)} \\
\bottomrule
\end{tabular}

\caption{Case study, binding: VideoRepick episode 8 (seed 7). Both
single-VLM pipelines pick a cube at the demonstration-time position;
\sys{}'s bank tracks the correct cube's two moves.}
\label{tab:case-vrp}
\end{table}

\begin{table}[tb]
\centering
\scriptsize
\setlength{\tabcolsep}{3pt}
\begin{tabular}{r >{\raggedright\arraybackslash}p{4.0cm} >{\raggedright\arraybackslash}p{4.0cm} >{\raggedright\arraybackslash}p{4.5cm}}
\toprule
step & live oracle (this run) & predicted subgoal & writer claim $\to$ verdict \\
\midrule
\multicolumn{4}{l}{\textbf{GroundSG+QwenVL}} \\
0 & move to the nearest right target by circling around the stick clockwise & move to the nearest right target by circling around the stick counterclockwise &  \\
\multicolumn{4}{l}{\emph{outcome:} failure (episode ends at step 48)} \\
\midrule
\multicolumn{4}{l}{\textbf{MemER}} \\
0 & move to the nearest right target by circling around the stick clockwise & move to the nearest right target by circling around the stick clockwise &  \\
48 & move to the nearest right target by circling around the stick counterclockwise & move to the nearest right target by circling around the stick clockwise &  \\
112 & move to the nearest left target by circling around the stick counterclockwise & move to the nearest left target by circling around the stick clockwise &  \\
\multicolumn{4}{l}{\emph{outcome:} failure (episode ends at step 160)} \\
\midrule
\multicolumn{4}{l}{\textbf{\sys{}}} \\
demo & \multicolumn{3}{p{12.7cm}}{bank after the demonstration replay: demo showed: stroke 1: move to the nearest right target by circling around the stick clockwise; demo showed: stroke 2: move to the nearest right target by circling around the stick counterclockwise; demo showed: stroke 3: move to the nearest left target by circling around the stick counterclockwise} \\
0 & move to the nearest right target by circling around the stick clockwise & move to the nearest right target by circling around the stick clockwise &  \\
48 & move to the nearest right target by circling around the stick counterclockwise & move to the nearest right target by circling around the stick counterclockwise & 1 move line $\to$ \textsc{admit} \\
96 & move to the nearest right target by circling around the stick counterclockwise & move to the nearest right target by circling around the stick counterclockwise & completed: move to the nearest right target by circling around the stick counterclockwise $\to$ \textsc{admit} \\
112 & move to the nearest left target by circling around the stick counterclockwise & move to the nearest left target by circling around the stick counterclockwise & completed: move to the nearest right target by circling around the stick counterclockwise $\to$ \textsc{admit} \\
\multicolumn{4}{l}{\emph{outcome:} success (episode ends at step 176)} \\
\bottomrule
\end{tabular}

\caption{Case study, direction: RouteStick episode 0 (seed 7). The
stroke sense is read at chance by the VLM pipelines and mechanically by
\sys{}'s reader; progress is advanced by admitted completion claims.}
\label{tab:case-rs}
\end{table}

\begin{figure}[tb]
\centering
\includegraphics[width=\linewidth]{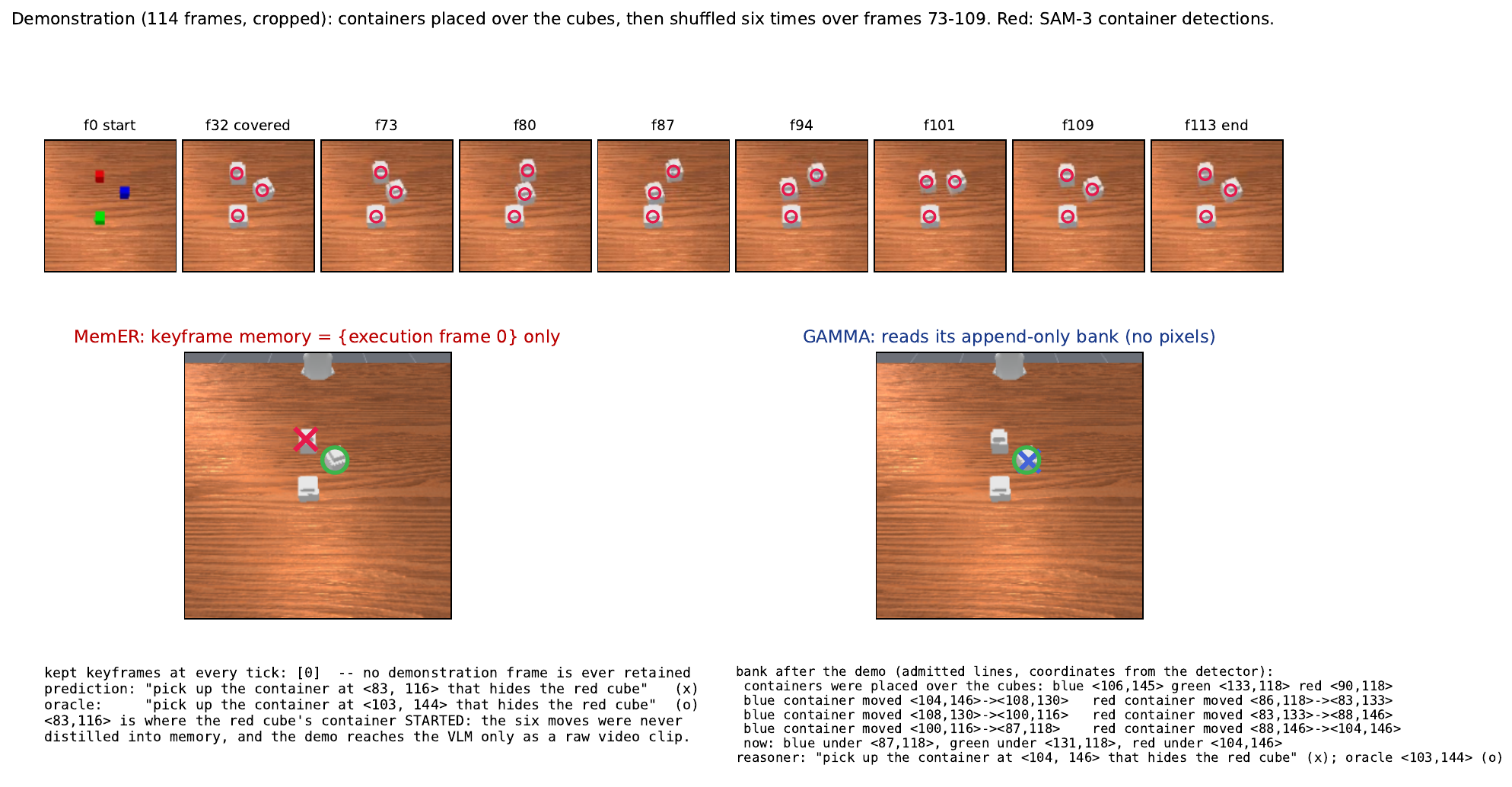}
\caption{A dropped decisive event, on one held-out VideoUnmaskSwap
episode (seed 7, episode 9) seen by both systems. \textbf{Top:} the
demonstration shuffles the three containers six times over frames
73--109 (SAM-3 container detections in red). \textbf{Bottom left:}
MemER's keyframe memory holds only execution frame 0 at every tick and
never a demonstration frame (the demonstration is attached to each query
as a raw video clip instead); its prediction points at the container
standing where the red cube's container \emph{started}, now hiding the
blue cube, 34\,px from the oracle. \textbf{Bottom right:} \sys{}'s bank
carries the six moves as individually admitted lines and the resulting
state; the reasoner's lookup lands within 2\,px of the oracle.}
\label{fig:memer-case}
\end{figure}

\section{Seed variance}
\label{app:seeds}

Table~\ref{tab:seeds} gives \sys{}'s per-task success for each of the
three serving seeds (identical agent checkpoints and configuration;
only the environment seed differs). Suite averages are
$66.7/66.3/65.2$---a $1.5$-point spread---and the per-family pattern
is stable across seeds.

\begin{table}[tb]
\centering
\scriptsize
\setlength{\tabcolsep}{3pt}
\resizebox{\textwidth}{!}{%
\begin{tabular}{l|cccc|cccc|cccc|cccc|c}
\toprule
Seed & PickX & BinF & SwingX & StopC & VU & BU & VUS & BUS
& PH & VRP & VPB & VPO & MC & IP & PL & RS & Avg \\
\midrule
7 & 100.0 & 63.3 & 66.7 & 43.3 & 86.7 & 83.3 & 80.0 & 36.7
& 70.0 & 76.7 & 73.3 & 70.0 & 60.0 & 6.7 & 96.7 & 53.3 & 66.7 \\
8 & 96.7 & 66.7 & 73.3 & 36.7 & 83.3 & 93.3 & 66.7 & 40.0
& 73.3 & 63.3 & 76.7 & 63.3 & 63.3 & 3.3 & 100.0 & 60.0 & 66.3 \\
9 & 100.0 & 83.3 & 66.7 & 26.7 & 90.0 & 76.7 & 80.0 & 30.0
& 70.0 & 56.7 & 76.7 & 66.7 & 63.3 & 0.0 & 100.0 & 56.7 & 65.2 \\
\midrule
mean & 98.9 & 71.1 & 68.9 & 35.5 & 86.7 & 84.4 & 75.6 & 35.6
& 71.1 & 65.6 & 75.5 & 66.7 & 62.2 & 3.3 & 98.9 & 56.7 & 66.0 \\
\bottomrule
\end{tabular}}
\caption{\sys{} per-task success (\%) by serving seed, 30
episodes/task each.}
\label{tab:seeds}
\end{table}

Table~\ref{tab:seeds-0p8b} gives the same breakdown for the 0.8B agents
(Section~\ref{sec:ablation}): suite averages are
$59.2/57.5/55.6$---a $3.6$-point spread---so the
scale comparison of Table~\ref{tab:ablation} rests on three seeds at both sizes.

\begin{table}[tb]
\centering
\scriptsize
\setlength{\tabcolsep}{3pt}
\resizebox{\textwidth}{!}{%
\begin{tabular}{l|cccc|cccc|cccc|cccc|c}
\toprule
Seed & PickX & BinF & SwingX & StopC & VU & BU & VUS & BUS
& PH & VRP & VPB & VPO & MC & IP & PL & RS & Avg \\
\midrule
7 & 73.3 & 46.7 & 63.3 & 46.7 & 66.7 & 76.7 & 46.7 & 30.0
& 53.3 & 73.3 & 90.0 & 76.7 & 46.7 & 0.0 & 96.7 & 60.0 & 59.2 \\
8 & 76.7 & 60.0 & 43.3 & 43.3 & 63.3 & 80.0 & 43.3 & 23.3
& 53.3 & 63.3 & 83.3 & 83.3 & 40.0 & 0.0 & 100.0 & 63.3 & 57.5 \\
9 & 80.0 & 56.7 & 50.0 & 30.0 & 66.7 & 73.3 & 40.0 & 23.3
& 46.7 & 70.0 & 90.0 & 73.3 & 36.7 & 0.0 & 100.0 & 53.3 & 55.6 \\
\midrule
mean & 76.7 & 54.4 & 52.2 & 40.0 & 65.6 & 76.7 & 43.3 & 25.6
& 51.1 & 68.9 & 87.8 & 77.8 & 41.1 & 0.0 & 98.9 & 58.9 & 57.4 \\
\bottomrule
\end{tabular}}
\caption{\sys{} with the 0.8B agents: per-task success (\%) by serving seed,
30 episodes/task each.}
\label{tab:seeds-0p8b}
\end{table}

\section{Configuration details}
\label{app:config}

\textbf{Models and finetuning.} Both agents share one Qwen3.5-9B base
with separate LoRA adapters: rank 16, $\alpha=32$, learning rate
$10^{-4}$, one epoch, effective batch 64.

\textbf{Corpus.} ${\sim}55$k writer records and matching reasoner
records over 1{,}600 recorded episodes; validation split by episode
index.

\textbf{Context geometry.} One tick $=$ one action chunk of $H=16$
control steps; $n=5$ frames per window (stride 4, boundaries included);
the writer sees the last 8 bank lines; the reasoner sees the full bank.

\textbf{Serving.} Two endpoints reproduce the training geometry:
\texttt{/reset} replays the demonstration prefix through the writer in
16-step windows to build the demo record; \texttt{/tick} runs
writer-then-reasoner once per 16 steps (Algorithm~\ref{alg:tick}).

\begin{algorithm}[tb]
\caption{One episode of \sys{}}
\label{alg:tick}
\footnotesize
\begin{algorithmic}[1]
\State $\bank \gets$ [initial-scene line from $D_0$] \Comment{tick 0, never \textsc{none}}
\State build demo record: replay the demonstration prefix through lines 4--6
\For{each tick $k$ of the execution phase}
  \State $D \gets \mathrm{SAM3}(o)$ on the window frames
  \State $\mathcal{C}_k \gets$ claims from $A_1(I, o, D, \bank, \sub_{k-1})$
  and $A_2(I, \bank)$
  \State $(\bank, \sub_k) \gets$ Algorithm~\ref{alg:harness} on
  $(\mathcal{C}_k, e_k)$ \Comment{admit / defer / correct / reject}
  \State execute $H$ steps of $\pi_{0.5}(\cdot, \sub_k)$
\EndFor
\end{algorithmic}
\end{algorithm}

\section{RoboMemArena details}
\label{app:rma}
\textbf{Transfer.} Nothing in \sys{} binds to RoboMME: the contracts
assume a detector, a tick grid, recorded episodes and an evidence stream
to verify claims against. The port changed two geometry constants (tick
$=10$ steps, $3$ frames per window), the detector prompts, and the
harness's evidence predicates, which here read the executor's
proprioceptive state: pick (fingers stalled inside the object's grasp
band, then the hand rises 3\,cm), place (carry, then release), pour
(wrist rotated $\geq 25^\circ$ from the carry orientation and back), open
and close (handle grasp and stroke). The agent code, the corpus
generator and the harness logic are those of Section~\ref{sec:method}.
Because every RoboMemArena task has a fixed subtask sequence, the
reasoner is pinned to the recorded plan, and the memory stage's output
is \emph{when} the executor is handed the next subtask. The reported
configuration runs the writer as on RoboMME: its completion claims are
verified against the harness's predicates and, like the harness's own
evidence-raised claims, admitted once the predicate has held for two
ticks. On the recorded episodes the predicates, with the grasp-width
priors used at deployment, fire inside their own segment in 879 of 888
cases. With the writer the system reaches 38.8 / 59.0; the harness's predicates alone, with no writer, give 38.7 / 58.6: with fixed plans the writer's only role is to raise completion claims, which the predicates can raise themselves, so the verified hand-off carries the gain. Admitting a verified writer claim on the tick it is raised instead gives 19.2 / 46.8 (10--20 ep.): the claims are correct but early, the hand-off effect described below.
\textbf{Executor.} The authors' bundled recipe: $\pi_{0.5}$ initialised
from the base checkpoint, action horizon 10, batch 128, 40k updates,
warm-up 10k then constant $5\times10^{-5}$, EMA 0.999, trained on the
2{,}600 released demonstrations (100 per task, segmented into subtasks)
with each segment's subtask text as the prompt. State normalisation
statistics match the released checkpoint's to three decimals and the
action quantiles differ by at most 0.04.
\textbf{Protocol.} Task success requires every scoring stage of an
episode; subtask success is the fraction of stages reached. Episodes
run 2{,}500 steps with 10 actions per policy call. The authors'
evaluation scripts default to seed 100 with one episode per task; the
released demonstrations are seeds 100 and above, so that default replays
training layouts. We evaluate on seeds 50--99 (50 episodes per task, 10
for the marked rows). On tasks 1--4 the authors' released checkpoint
scores the same on training and held-out seeds (70/100/90/0 vs.\
70/90/100/0 task success), so the choice of seeds is not what separates
the rows of Table~\ref{tab:rma-full}.
\textbf{Hand-off timing.} The released segments end at completion:
place segments end with the gripper fully open and the arm retracted
($100\%$ of segments), pour segments after the wrist has returned
($|\Delta\text{rot}|=0.01$\,rad), pick segments with the fingers closed
on the object. The benchmark's scoring predicates fire earlier (object
entering the receptacle while still grasped; tilt peak), so a plan
advanced on them hands the executor a subtask mid-motion; advancing on
the segment-completion conditions instead (object released, wrist
returned) removes part of the gap (19.2 task success), and the
harness's verified transitions, which also wait two ticks after the
predicate holds, remove most of it (38.7).
\textbf{Two executors, complementary failures.} The authors' released
$\pi_{0.5}$ under our protocol reaches 32.7 / 55.2 with the
benchmark-predicate feed and 31.5 / 56.1 with the harness's
timing: it is stronger on the microwave and cabinet-transfer tasks and
weaker on pouring. Both executors, on held-out and on training layouts,
fail the drawer tasks (4, 5, 11, 14); on the three-drawer tasks 4 and 5
ours opens and closes the first drawer in every episode and then fails
the second, while theirs rarely opens the first.

\begin{table}[tb]
\centering
\scriptsize
\setlength{\tabcolsep}{3pt}
\resizebox{\textwidth}{!}{%
\begin{tabular}{l|cccc|c}
\toprule
Row (task / subtask success, \%) & Transferring & Occlusion & Counting & Sequence & Average \\
\midrule
\multicolumn{6}{l}{\emph{Our protocol: held-out layouts (seeds 50--99), 50 episodes per task unless marked}} \\
$\pi_{0.5}$ executor, plan fed on benchmark stage predicates (privileged) & 3.5 / 23.6 & 2.4 / 40.7 & 11.4 / 36.5 & 35.0 / 68.1 & 10.0 / 41.1 \\
$\pi_{0.5}$ executor, plan fed on segment-completion conditions (privileged, 10 ep.) & 7.5 / 24.6 & 4.5 / 41.9 & 22.9 / 43.3 & 65.0 / 83.8 & 19.2 / 46.1 \\
\rowcolor{oursrow} \sys{} full pipeline: writer + harness, claims settled two ticks & 27.4 / 37.6 & 19.2 / 50.1 & 52.3 / 67.5 & 80.7 / 90.3 & 38.8 / 59.0 \\
\rowcolor{oursrowlite} \sys{} harness predicates only, no writer & 27.5 / 37.3 & 18.9 / 49.2 & 52.0 / 67.2 & 81.0 / 90.7 & 38.7 / 58.6 \\
\sys{} writer + harness, claims admitted on the tick they are verified (10--20 ep.) & 1.2 / 20.2 & 3.2 / 45.1 & 24.3 / 42.1 & 72.5 / 86.2 & 19.2 / 46.8 \\
authors' released $\pi_{0.5}$, plan fed on benchmark stage predicates (10 ep.) & 52.5 / 62.5 & 17.3 / 40.4 & 12.9 / 53.8 & 90.0 / 91.2 & 32.7 / 55.2 \\
authors' released $\pi_{0.5}$, \sys{} harness timing (10 ep.) & 50.0 / 62.9 & 20.9 / 45.8 & 5.7 / 47.1 & 87.5 / 93.3 & 31.5 / 56.1 \\
\midrule
\multicolumn{6}{l}{\emph{Authors' protocol, as reported}} \\
authors' reported $\pi_{0.5}$ & 20.0 / 42.8 & 12.7 / 17.2 & 14.3 / 50.9 & 60.0 / 71.6 & 21.5 / 38.7 \\
authors' reported PrediMem (full system) & 22.5 / 45.2 & 27.3 / 38.4 & 45.7 / 69.3 & 72.5 / 89.5 & 38.5 / 55.2 \\
authors' reported ground-truth subtask feed & 32.5 / 54.8 & 33.6 / 49.8 & 51.4 / 75.6 & 85.0 / 92.3 & 46.1 / 64.8 \\
\bottomrule
\end{tabular}}
\caption{RoboMemArena by task family (task / subtask success, \%). Family
sizes: sequence 4 tasks, counting 7, occlusion 11, transferring 4. The
upper block is our protocol; the lower block reproduces the authors'
Table~2 under their own protocol for reference.}
\label{tab:rma-full}
\end{table}

\section{Bank quality and additional material}
\label{app:bankquality}
\textbf{Corpus.} The released corpus (Qwen3.5-9B configuration) holds
$54{,}829$ training and $5{,}904$ validation writer records over
$1{,}600$ episodes; $11{,}506$ training records carry an event target
(initial-scene $2{,}056$, demo-record $2{,}543$, move-chain $1{,}901$,
completion $3{,}468$, other observation $1{,}538$) and the remaining
$43{,}323$ are \textsc{none}. Two audits run before training.
\emph{Knowability}: $97.5\%$ of training-target coordinates
($97.3\%$ validation) lie within $2$\,px of a detection in the
writer's window or a coordinate in the bank lines it sees; the residue
are chain coordinates carried from lines outside the writer's 8-line
context. \emph{Evidence gate}: $94.5\%$ of training move lines
($96.2\%$ validation) carry a measured displacement of at least
$8$\,px between their detected endpoints.

\textbf{Open-loop agent quality.} Table~\ref{tab:openloop} scores the
released 9B agents on the $5{,}904$ held-out validation ticks. The writer
is rolled out autoregressively on its own bank (not the ground-truth
bank); \emph{exact} is per-tick string match of the line (or
\textsc{none}) with the target after stripping the detection suffix,
\emph{recall} the share of event ticks whose line is reproduced
exactly, \emph{precision} the share of written lines that fall on an
event tick, and \emph{known} the share of coordinates the writer cites
that appear in its inputs. The reasoner is scored on the writer's
predicted banks against its training target (the oracle string with
bank-snapped coordinates and the phase conventions).

\begin{table}[tb]
\centering
\scriptsize
\setlength{\tabcolsep}{4pt}
\begin{tabular}{l|rrrr|rr}
\toprule
& \multicolumn{4}{c|}{Writer (autoregressive rollout)} & \multicolumn{2}{c}{Reasoner} \\
Task & exact\% & recall\% & precision\% & known\% & ticks & exact\% \\
\midrule
PickXtimes       & 99.2 & 97.6 & 97.6 & 97.5 & 260 & 80.0 \\
BinFill          & 94.4 & 76.4 & 98.6 & 94.3 & 396 & 68.2 \\
SwingXtimes      & 95.4 & 87.0 & 95.3 & 99.1 & 262 & 94.7 \\
StopCube         & 80.0 & 65.2 & 91.9 & 96.3 & 240 & 88.7 \\
VideoUnmask      & 100.0 & 100.0 & 100.0 & 100.0 & 126 & 100.0 \\
ButtonUnmask     & 96.2 & 85.7 & 96.9 & 95.5 & 159 & 98.1 \\
VideoUnmaskSwap  & 94.4 & 88.5 & 98.8 & 99.7 & 198 & 92.9 \\
ButtonUnmaskSwap & 91.2 & 77.1 & 97.5 & 97.5 & 239 & 80.8 \\
PickHighlight    & 89.7 & 64.9 & 95.7 & 94.7 & 213 & 76.1 \\
VideoRepick      & 83.6 & 50.4 & 90.3 & 95.5 & 494 & 81.4 \\
VideoPlaceButton & 96.2 & 73.0 & 84.8 & 93.6 & 1092 & 98.6 \\
VideoPlaceOrder  & 96.0 & 68.3 & 88.2 & 94.4 & 1313 & 97.7 \\
MoveCube         & 93.2 & 70.5 & 88.1 & 90.8 & 263 & 88.6 \\
InsertPeg        & 92.4 & 58.3 & 91.3 & 96.6 & 316 & 71.5 \\
PatternLock      & 59.3 & 38.7 & 87.1 & 95.1 & 113 & 58.4 \\
RouteStick       & 64.1 & 34.0 & 82.1 & 86.2 & 220 & 72.7 \\
\midrule
All              & 91.8 & 67.8 & 91.7 & 95.2 & 5904 & 88.2 \\
\bottomrule
\end{tabular}
\caption{Open-loop quality of the released agents on the held-out
validation split ($5{,}904$ ticks, $1{,}211$ event ticks). The
direction tasks are where the writer's own lines are weakest, which is
why their demonstration records are reader-owned at deployment
(Appendix~\ref{app:motion}).}
\label{tab:openloop}
\end{table}

\textbf{Detector.} SAM-3 (\texttt{facebook/sam3}) is run with the text
prompts of Table~\ref{tab:detector}, score threshold $0.35$ and mask
threshold $0.5$; each detection is reported as its mask centroid on the
$256$-px agent-view frame in the benchmark oracle's own
$\langle \text{row}, \text{column}\rangle$ order (written
$\langle x, y\rangle$ generically in the main text), and $14$\,px is
the hit radius used by the corpus audit and the wrong-instance test.
The same prompts and thresholds are used at corpus construction and at
deployment.

\begin{table}[tb]
\centering
\footnotesize
\setlength{\tabcolsep}{6pt}
\begin{tabular}{ll|ll}
\toprule
Class & Text prompt & Class & Text prompt \\
\midrule
red\_cube   & a small red block on the table   & target & a flat square target marker on the table \\
green\_cube & a small green block on the table & bin    & a white open box \\
blue\_cube  & a small blue block on the table  & peg    & a long cylindrical peg on the table \\
container   & white container box              & stick  & a thin stick standing on the table \\
button      & a grey push button on the table  & arm    & robot arm \\
\bottomrule
\end{tabular}
\caption{SAM-3 text prompts per object class (score threshold $0.35$).}
\label{tab:detector}
\end{table}

\textbf{Prompt templates.} Both agents use a fixed system prompt and a
templated user turn; the writer's turn interleaves the window frames
with their detection strings.

{\scriptsize
\begin{verbatim}
[writer, system] You are the robot's memory writer (agent-1). Each tick you watch the
last 16 control steps (sampled frames with per-frame object detections), read the
memory bank and the planner's expected next subgoal, and write ONE line faithfully
describing what happened in this window with grounded object coordinates, in the
format '[event] ...  [sam] name<x, y>'. Write one line per event; if two things
happened, write two lines. If nothing new happened, write NONE.

[writer, user]  Task: <instruction>
                Phase: <demo | execution>
                Expected next subgoal from the planner: <previous reasoner output>
                Memory bank so far:            <last 8 admitted lines, or (empty)>
                Current window frames and detections:
                <image> +0: name<x, y> name<x, y> ...
                <image> +4: ...      (5 frames, stride 4)
                Write the memory line for this window (or NONE).

[reasoner, system] You are the robot's planner (agent-2). You read the task and the
robot's memory bank of grounded events, and you output the robot's current subgoal
for this control chunk. Judge progress from the bank's completed lines. Answer with
the subgoal sentence only, lowercase, no trailing period, copying object coordinates
from the memory bank in the form <x, y>. During the demo phase answer exactly: hold
and wait for the demo to complete.

[reasoner, user] Task: <instruction>
                 Memory bank:                  <all admitted lines>
                 It is now the <DEMO | EXECUTION> phase.
                 What is the robot's current subgoal?
\end{verbatim}}

\textbf{Bank walkthroughs.} The four banks below are verbatim from
successful held-out episodes of the released 9B configuration (serving
seed 7), one per memory family, with the detection suffixes elided for
width. They show what the reasoner actually reads: in every case the
subgoal it must emit is a lookup or a count over these lines, never an
inference from pixels. Each bank is followed by the agent-view frames at
the ticks where its decisive lines were admitted, with the coordinates
those lines cite drawn on the frame, so every line can be checked
against the evidence that admitted it.

\emph{Temporal} (PickXtimes, ep 3). Counting is carried by the
completion lines alone: the reasoner emits the $n$-th pick by counting
admitted \emph{completed: pick up} lines, and each is admitted only when
the window shows the lift.

{\scriptsize
\begin{verbatim}
instruction: pick up the blue cube and place it on the target, repeating this
             action four times, then press the button to stop
[event] initial scene: green cube at <94,110>, red cube at <88,142>,
        button at <61,141>, target at <61,106>, blue cube at <71,170>
[event] completed: pick up the blue cube at <71,170> for the first time
[event] completed: place the blue cube onto the target at <61,106>
[event] completed: pick up the blue cube at <58,102> for the second time
[event] completed: place the blue cube onto the target at <61,106>
[event] completed: pick up the blue cube at <57,102> for the third time
[event] completed: place the blue cube onto the target at <61,106>
[event] completed: pick up the blue cube at <57,102> for the fourth time
[event] completed: place the blue cube onto the target at <61,106>
\end{verbatim}}

\begin{figure}[tb]
\centering
\includegraphics[width=\linewidth]{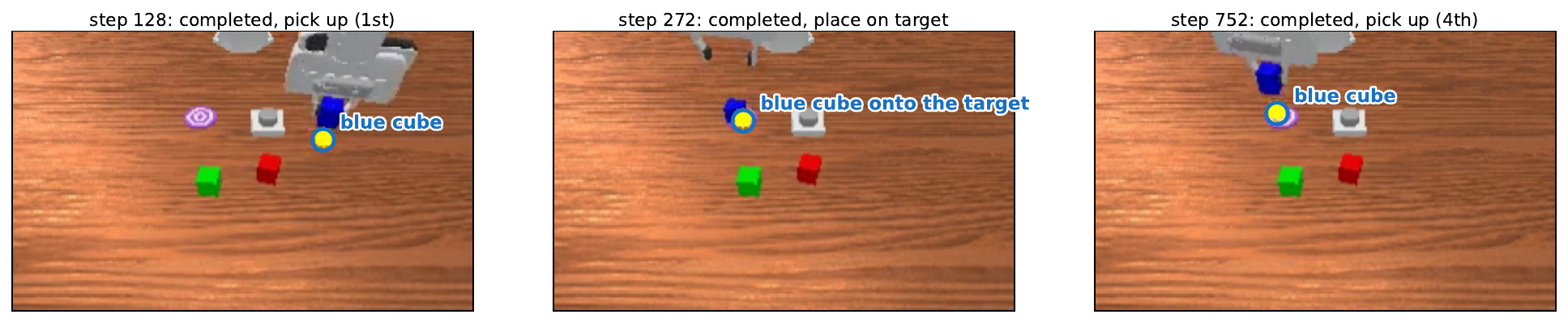}
\caption{Frames behind the PickXtimes bank above (episode 3, agent view; markers
are the coordinates the admitted line cites). Each \emph{completed: pick
up} line is admitted at the tick that shows the cube leaving the table in
the gripper, and each \emph{place} line at the tick the cube reaches the
target; counting the fourth pick is then a count over admitted lines.}
\label{fig:bank-pickx}
\end{figure}

\emph{Spatial} (ButtonUnmaskSwap, ep 1; the case study of
Table~\ref{tab:case-bus}). Two memories coexist: the shuffle as a chain
of per-move lines closed by a state summary, and the press count as
completion lines. The reasoner's pick-up cites $\langle 110,85\rangle$
by reading the summary, not by re-deriving the swaps.

{\scriptsize
\begin{verbatim}
instruction: first press both buttons on the table, then pick up the container
             hiding the red cube, finally pick up another container hiding the
             green cube
[event] initial scene: green cube at <88,166>, blue cube at <112,86>,
        red cube at <102,124>, button at <66,155>, button at <63,104>
[event] observed: containers were placed over the cubes - the blue cube at
        <112,86>, the green cube at <88,166>, the red cube at <102,124>
[event] observed: the blue cube container moved from <96,92> to <90,108>
[event] observed: the red cube container moved from <114,118> to <120,100>
[event] observed: the blue cube container moved from <92,110> to <98,121>
[event] observed: the red cube container moved from <119,99> to <112,86>
[event] observed: now the blue cube is under the container at <98,121>, the
        green cube is under the container at <86,167>, the red cube is under
        the container at <112,86>
[event] completed: press the first button at <66,155>
[event] completed: press the second button at <63,104>
[event] completed: pick up the container at <110,85> that hides the red cube
\end{verbatim}}

\begin{figure}[tb]
\centering
\includegraphics[width=\linewidth]{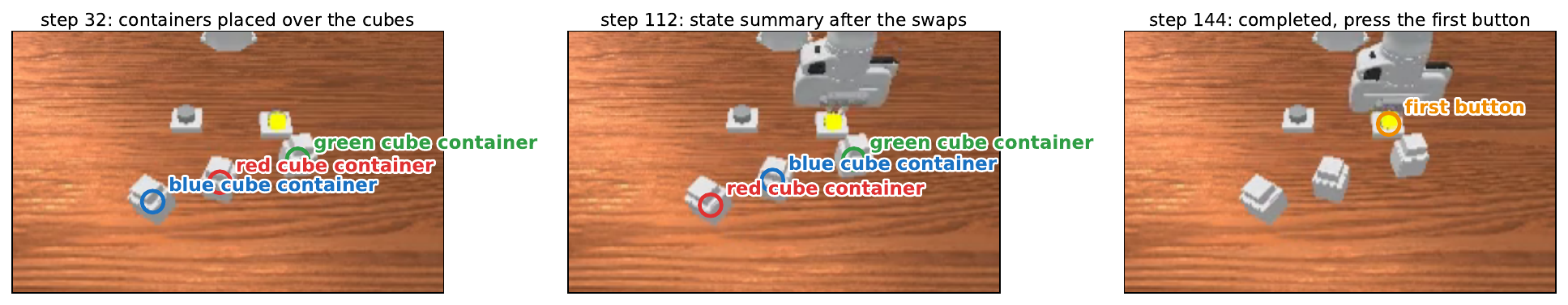}
\caption{Frames behind the ButtonUnmaskSwap bank above (episode 1).
\textbf{Left:} the containers are admitted once they are detected over
the cubes. \textbf{Middle:} after the swaps, the summary line records
which container hides which cube, and its three coordinates are the
detections in this window. \textbf{Right:} the press completion is
admitted at the tick the button is occluded under the gripper---the
evidence the \textsc{defer} gate waits for.}
\label{fig:bank-bus}
\end{figure}

\emph{Object} (VideoRepick, ep 1). The demonstration names the cube
once; the tracker lines carry its identity through the shuffle to a
final summary, so at execution the reasoner cites
$\langle 135,114\rangle$ rather than the demonstration-time position
$\langle 106,137\rangle$ that both single-VLM baselines cite
(Table~\ref{tab:case-vrp}).

{\scriptsize
\begin{verbatim}
instruction: watch the video carefully, then pick up the same block that was
             previously picked up again, finally put it down and press the
             button to stop
[event] initial scene: red cube at <135,114>, red cube at <106,136>,
        red cube at <78,138>, button at <59,121>, container at <14,128>
[event] demo showed: pick up the cube at <106,137>
[event] demo showed: drop the cube on the table
[event] observed: the correct cube moved from <106,137> to <122,145>
[event] observed: the correct cube moved from <122,145> to <133,137>
[event] observed: the correct cube moved from <135,131> to <135,114>
[event] observed: the cube picked in the video is now at <135,114>
        - this is the correct cube
[event] execution phase begins - scene: red cube at <135,114>, red cube at
        <78,138>, red cube at <105,137>, button at <59,121>
[event] completed: pick up the correct cube at <135,114> for the first time
[event] completed: put it down at <105,137>
\end{verbatim}}

\begin{figure}[tb]
\centering
\includegraphics[width=\linewidth]{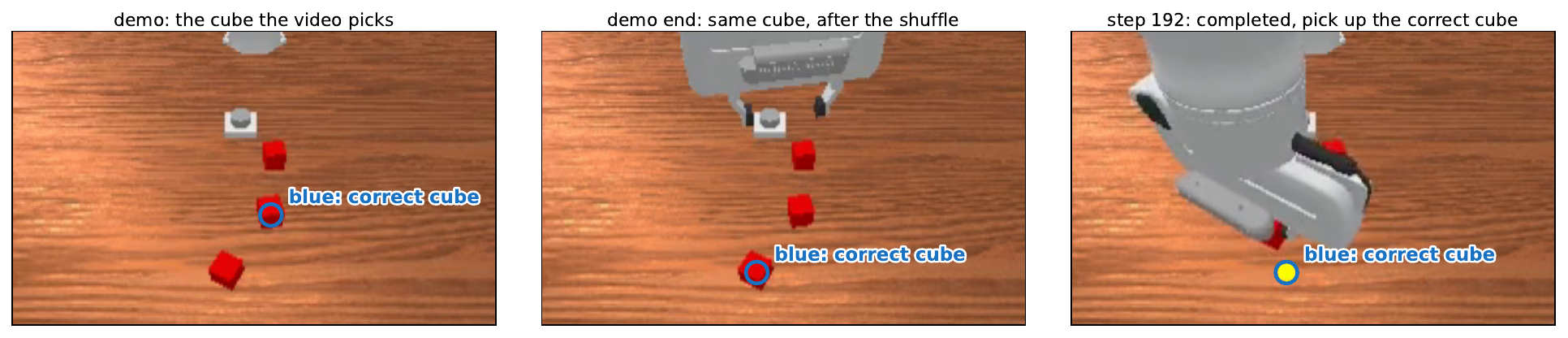}
\caption{Frames behind the VideoRepick bank above (episode 1); the marked cube is
the same physical object in all three. \textbf{Left:} the cube the
demonstration picks, at $\langle 106,136\rangle$. \textbf{Middle:} the
end of the demonstration, after the shuffle has moved it to
$\langle 135,114\rangle$, the coordinate the tracker lines carry.
\textbf{Right:} the admitted pick at execution, on that coordinate
rather than on the demonstration-time one.}
\label{fig:bank-vrp}
\end{figure}

\emph{Procedural} (RouteStick, ep 13). The numbered \emph{demo showed}
lines are the reader's, not the writer's (Appendix~\ref{app:motion});
the arc-tip lines around them are the writer's own observations, and
stroke progress is advanced by the admitted completion.

{\scriptsize
\begin{verbatim}
instruction: watch the video carefully, then use the stick attached to the
             robot to navigate around the sticks on the table, following the
             same path
[event] initial scene: blue cube at <72,119>, red cube at <69,76>, green cube
        at <64,86>, arm at <6,49>, peg at <80,155>, peg at <90,194>
[event] observed: the arc tip moved from <46,49> to <45,69>
[event] observed: the arm is at <33,97> and begins a new stroke
[event] observed: the arc tip moved from <85,84> to <78,90>
[event] demo showed: stroke 1: move to the nearest left target by circling
        around the stick counterclockwise
[event] observed: the arc tip moved from <84,91> to <84,96>
[event] demo showed: stroke 2: move to the nearest left target by circling
        around the stick clockwise
[event] execution phase begins - scene: blue cube at <72,119>, green cube at
        <64,86>, peg at <80,155>, peg at <90,194>
[event] observed: the arc tip moved from <44,45> to <48,56>
[event] completed: move to the nearest left target by circling around the
        stick counterclockwise
\end{verbatim}}

\begin{figure}[tb]
\centering
\includegraphics[width=\linewidth]{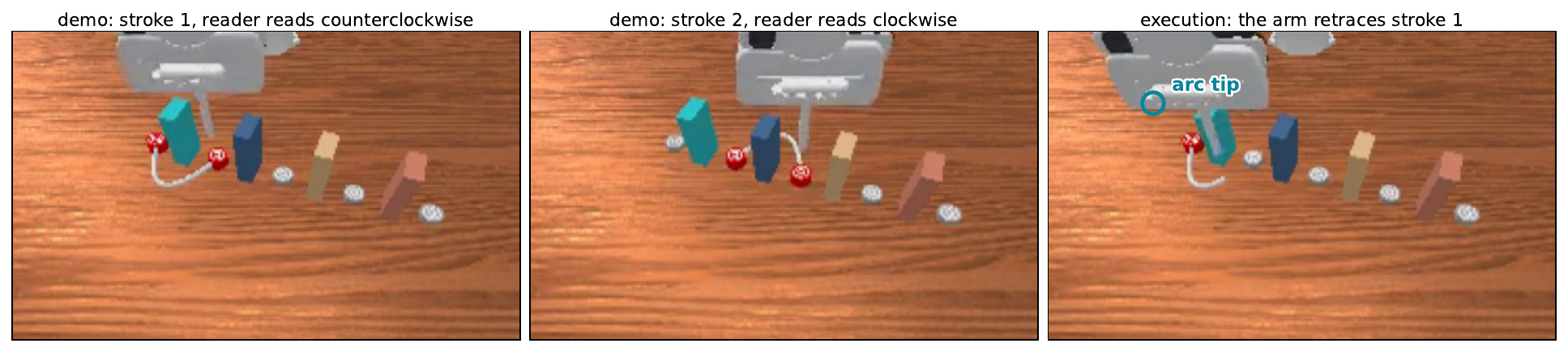}
\caption{Frames behind the RouteStick bank above (episode 13). \textbf{Left} and
\textbf{middle:} the two demonstration strokes whose turning sense the
deterministic reader extracts from the ink the stick leaves (the pale
arc), and on which the writer's own words measured at chance.
\textbf{Right:} the arm retracing stroke 1 at execution, with the
arc-tip detection the writer records.}
\label{fig:bank-rs}
\end{figure}

Across the logged episodes of the released configuration (three serving
seeds) a bank
holds a median of $8$ lines (5th--95th percentile $3$--$20$, maximum
$73$) against a median episode of $240$ control steps (95th percentile
$1{,}296$). That compression is what lets the reasoner read the whole
history at every tick.

\end{document}